\pdfoutput=1
\documentclass[11pt]{article}

\usepackage[preprint]{acl}

\usepackage{times}
\usepackage{latexsym}

\usepackage[T1]{fontenc}

\usepackage[utf8]{inputenc}

\usepackage{microtype}

\usepackage{inconsolata}

\usepackage{graphicx}
\usepackage{booktabs}

\usepackage[table]{xcolor}
\usepackage{multirow}
\usepackage{subcaption}
\usepackage{tabularx}

\newcommand{\ws}[2]{%
    \ifnum#1=0
        \multicolumn{1}{c}{#2}%
    \else
        \multicolumn{1}{c}{\cellcolor{cyan!#1}{#2}}%
    \fi
}

\newcommand{\dcell}[2]{%
    \multicolumn{1}{c}{\cellcolor{cyan!#1}{#2}}%
}

\usepackage[most]{tcolorbox}

\newtcolorbox{promptbox}{
  colback=gray!5,
  colframe=gray!60,
  boxrule=0.5pt,
  arc=2pt,
  left=6pt,
  right=6pt,
  top=6pt,
  bottom=6pt
}

\usepackage{tikz}
\usetikzlibrary{positioning,calc}
\usepackage{graphicx}
\usepackage{xcolor}
\usepackage{adjustbox}

\title{Instructions for *ACL Proceedings}
\title{In the Eye of the Beholder:\\  Human Label Variation in Textual and Visual Entailment}
 
\newcommand{\projname}{\textsc{VariNLI-VE}}

\author{
  \textbf{Monica Riedler\textsuperscript{1,2}} \quad
  \textbf{Jasmin Orth\textsuperscript{1}} \quad
  \textbf{Raffaella Bernardi\textsuperscript{3}} \quad
  \textbf{Barbara Plank\textsuperscript{1,2}}
  \\
  \textsuperscript{1}MaiNLP, Center for Information and Language Processing, LMU Munich, Germany \\
  \textsuperscript{2}Munich Center for Machine Learning (MCML), Munich, Germany \\
  \textsuperscript{3}Free University of Bozen-Bolzano, Italy \\
  \small{
  \texttt{\{monica.riedler,b.plank\}@lmu.de} \quad
  \texttt{jasmin.orth@campus.lmu.de} \quad
  \texttt{raffaella.bernardi@unibz.it}
  }
}

\begin{document}
\maketitle
\begin{abstract}
Human Label Variation (HLV) is increasingly recognised as informative signal that reflects meaningful differences in how people interpret the same input. While HLV has been studied extensively in Natural Language Inference (NLI), much less is known about HLV in Visual Entailment (VE; \citealp{xie2019visualentailmentnoveltask}), where the textual premise is replaced by an image, keeping the hypothesis and label space fixed. We introduce \projname{}, a dataset for studying how HLV changes between paired NLI and VE instances. We re-annotate a subset of originally neutral SNLI \citep{bowman-etal-2015-large} instances and their visual counterparts, with the same annotators evaluating both the textual and visual versions. Each annotation includes a label, confidence score, and explanation, yielding 2,250 annotations. Labels and explanations are collected jointly, providing ecologically valid explanations. Because the image may reveal information not stated in the text, one might expect disagreement to decrease in VE. Instead, we find no clear overall reduction: disagreement decreases for some instances while it increases for others. Explanations reveal further variation, as annotators can retain the same label while changing their rationale. When evaluating vision-language models (VLMs) on the paired data, we find lower accuracy, weaker alignment with human label distributions, and generally higher uncertainty in VE than in NLI.
\end{abstract}

\section{Introduction}

\begin{figure}[t]
    \centering
    \includegraphics[width=\columnwidth]{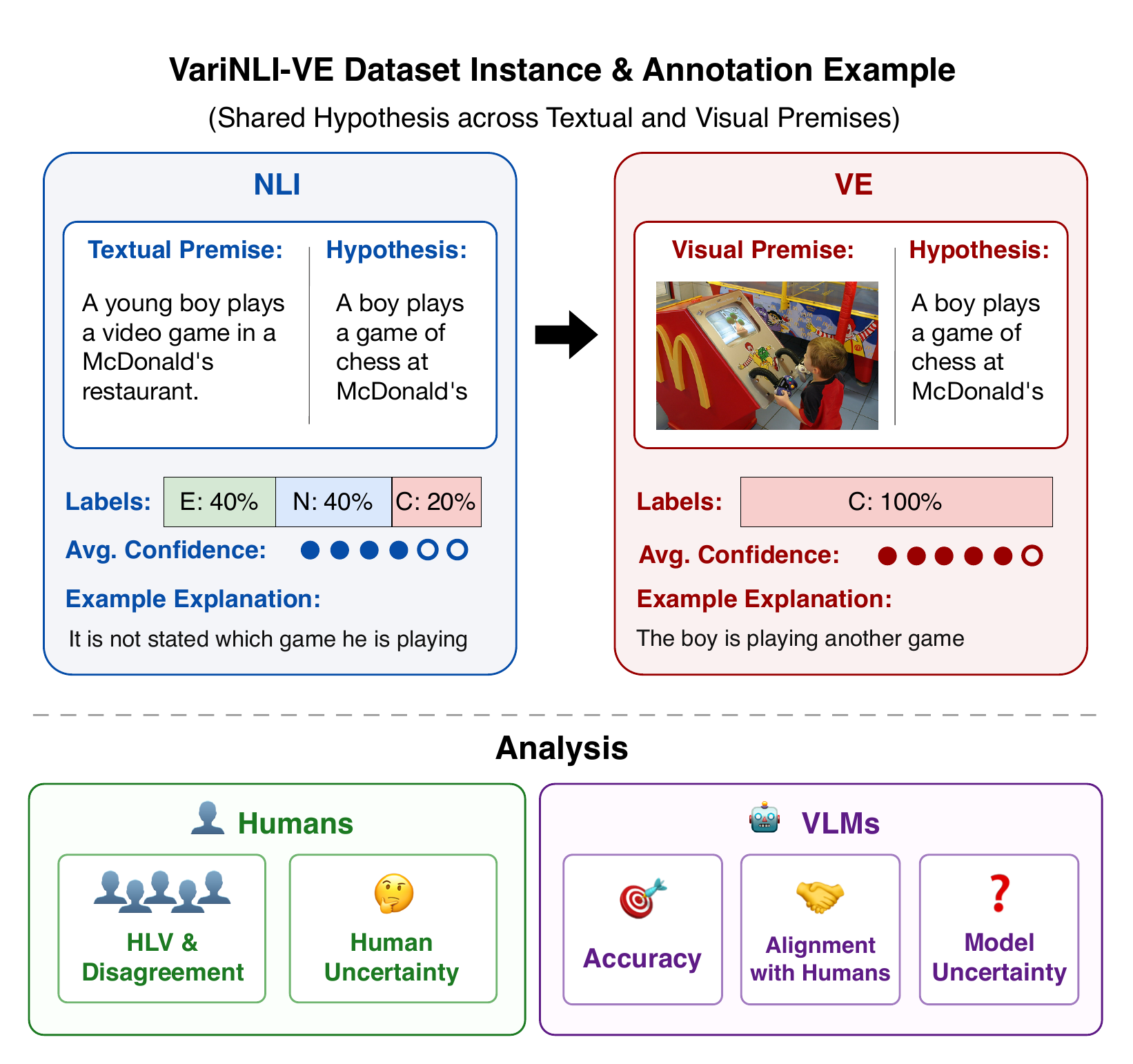}
    \caption{Setup and analyses. For each VariNLI-VE instance, we collect five annotations, including a label (E, N, C), confidence score (1-6), and explanation from the same annotators across NLI and VE tasks, and compare with model predictions and uncertainty estimates.}
    \label{fig:nli_ve_overview}
\end{figure}

Many Natural Language Processing tasks assume that each input is associated with a single correct label. In practice, however, annotators often disagree, giving rise to Human Label Variation (HLV; \citealp{plank-2022-problem}). Rather than being noise, such variation can reflect ambiguity, different interpretations, and multiple valid readings \citep{aroyo2015truthlie, uma-2022-survey}. While substantial prior work has studied variation and disagreement in the Natural Language Inference (NLI) task (\citealp{pavlick-kwiatkowski-2019-inherent,
nie-etal-2020-learn,
jiang-etal-2023-ecologically}, \textit{inter alia}), human inference is not restricted to textual evidence. In the Visual Entailment (VE) task \citep{xie2019visualentailmentnoveltask}, the premise is presented as an image rather than text, while the hypothesis and label space remain unchanged. However, HLV in VE remains underexplored. We therefore use VE as a test bed to study how HLV changes from NLI to VE when the same annotators judge the same hypothesis under textual and visual premises.

This comparison is particularly interesting because an image may provide evidence that is left unstated in a textual premise. Such additional evidence could resolve potential underspecification in the text, narrow the range of plausible interpretations, and reduce disagreement. Figure~\ref{fig:nli_ve_overview} illustrates such a case: the textual premise elicits disagreement among annotators, while evidence available in the corresponding image leads all annotators to unanimously assign the contradiction label.

To study this evidence shift in a paired setting, we introduce \projname, a dataset of 2,250 new annotations across NLI and VE, constructed from originally neutral SNLI instances \citep{bowman-etal-2015-large}. Since SNLI premises originate from image captions, each textual premise can be replaced with its corresponding source image. Each version of an NLI--VE pair receives five annotations consisting of a label, a natural-language explanation, and a self-reported confidence score. Crucially, the same five annotators judge both the textual and visual versions, enabling direct comparison across the two settings. Since labels and explanations are collected jointly, the dataset provides ecologically valid explanations \citep{jiang-etal-2023-ecologically} that are tied directly to the judgments they explain. To the best of our knowledge, \projname{} is the first dataset to provide multiple annotations, including explanations, from the same individuals for paired NLI and VE instances.

We further ask whether vision-language models (VLMs) reflect the human disagreement and uncertainty observed across NLI and VE. We evaluate a range of VLMs on the same paired instances, comparing their accuracy and predictive distributions with human judgments, and their uncertainty with human disagreement and reported confidence.

We make the following contributions:

\begin{itemize}
\item We release \projname, a dataset of 2,250 paired NLI--VE annotations for originally neutral SNLI instances, including labels, explanations, and confidence scores from the same annotators across textual and visual premises.\footnote{Data and code will be made publicly available with the camera-ready version. See additional examples in App. \ref{sec:appendix_examples}.}
\item Using this paired setup, we analyze how human label distributions, explanations, and confidence change between NLI and VE.
\item We evaluate how VLM predictions and uncertainty change across the paired tasks, and to what extent they align with human judgments.
\end{itemize}

Contrary to the expectation that visual evidence would reduce disagreement, we find no clear overall decrease from NLI to VE. Instead, disagreement decreases for some instances and increases for others. Human confidence remains generally high in both settings, suggesting that disagreement can reflect confidently held competing interpretations. Explanations add another layer to the HLV analysis: even when an annotator assigns the same label in NLI and VE, the rationale for that judgment can change, revealing variation that label distributions alone do not capture. For models, we find consistently lower accuracy and weaker alignment with human judgment distributions in VE compared to NLI, while model uncertainty is generally higher and only weakly tracks human uncertainty.

\section{Related Work}

\subsection{Human Label Variation \& NLI}

HLV \citep{plank-2022-problem} arises when multiple annotators assign different labels to the same instance. Rather than treating such disagreement as noise, prior work argues that it can reflect meaningful variation in human judgments due to ambiguity, plausible differences in interpretation, or annotator-specific factors \citep{aroyo2015truthlie, pavlick-kwiatkowski-2019-inherent, jiang-marneffe-2022-investigating}.
This has motivated work that preserves disagreement, as models trained on such data better reflect the range of human interpretations \citep{uma-2022-survey, fornaciari-etal-2021-beyond}.
The ChaosNLI dataset \citep{nie-etal-2020-learn} shows that HLV is widespread in NLI by collecting annotations from a large number of annotators.
Explanations can further reveal reasoning behind disagreement: the 
e-SNLI dataset \citep{esnli-2018-camburu} adds natural language explanations to SNLI \citep{bowman-etal-2015-large}, but these were collected post hoc and from annotators who differ from those providing the labels.
LiveNLI \citep{jiang-etal-2023-ecologically} and VariErr NLI \citep{weber-genzel-etal-2024-varierr} instead collect ecologically valid explanations, where annotators provide labels and explanations jointly, enabling variation to be studied beyond labels alone.

\subsection{VQA \& Answer Diversity}

Annotator variation and diversity also arise in Visual Question Answering (VQA), where question-image pairs may produce different responses due to ambiguity, subjectivity, or difficulty~\citep{gurari2017crowdverge, kafle2017visual, yang2018visual}.
Several VQA datasets collect multiple human answers per instance~\citep{antol2015vqa, goyal2017making, gurari2018vizwiz, marino2019ok}.
Existing approaches have investigated this variation through eye-tracking, linking disagreement to different gaze patterns~\citep{hindennach2024explaining} or through correlations between VLMs and human answer distributions~\citep{lan2025mind}, finding that even the best-performing models struggle to approximate human distributions.
While these prior VQA works establish disagreement and diversity as important phenomena in multimodal tasks, we study HLV across paired textual and visual evidence, using the same annotators in both settings.

\subsection{From NLI to Visual Entailment}

VE extends NLI to a multimodal setting by replacing the textual premise with an image while keeping the hypothesis and label space fixed \citep{xie2019visualentailmentnoveltask}.\footnote{This contrasts with the setup from \citet{vu-etal-2018-grounded}, where images are added to the premise rather than replacing it.}
This is possible for the SNLI dataset \citep{bowman-etal-2015-large}, since its premises are derived from Flickr30k image captions \citep{young-etal-2014-flickr30k} and can therefore be replaced with the corresponding images.
The e-SNLI-VE dataset \citep{Kayser_2021_ICCV} further adds natural language explanations from e-SNLI \citep{esnli-2018-camburu}. 
At the same time, replacing captions with images can change the information available to annotators, and \citet{Kayser_2021_ICCV} address resulting label mismatches by re-annotating neutral examples of the e-SNLI-VE development and test sets.
These resources make VE a natural setting for studying HLV under a caption-to-image shift. Yet HLV is not a primary focus of existing VE resources, and explanations are only partially ecologically valid. While re-annotated neutral instances receive new explanations and labels, most explanations are inherited from post-hoc explanations from e-SNLI. Thus, existing resources do not provide multiple labels and explanations for paired NLI-VE instances annotated by the same individuals. Our work is a first step towards addressing this gap.

\begin{figure}[t]
    \centering
    \includegraphics[width=\columnwidth]{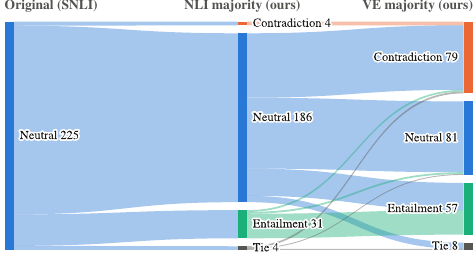}
    \caption{
    Label transitions from the original SNLI label to our newly derived NLI and VE majority labels in VariNLI-VE. Ties indicate cases with no majority label among five annotators. 
    }
    \label{fig:nli_ve_shift}
\end{figure}

\section{\projname}

We construct \projname{} from 225 SNLI instances \citep{bowman-etal-2015-large} and their source images. Each instance yields two annotation items that share the same hypothesis: an NLI item with the original textual premise and a VE item with the corresponding image. Each version receives five annotations consisting of a label, a natural-language explanation, and a self-reported confidence score. Crucially, the same five annotators judge both the textual and visual versions of each instance, yielding 2,250 annotations in total (225 instances $\times$ 2 versions $\times$ 5 annotators). Confidence is expressed on a six-point scale, from 1 (lowest) to 6 (highest). Labels, explanations, and confidence scores are collected jointly, making the explanations ecologically valid \citep{jiang-etal-2023-ecologically}. Item order is randomized during annotation, while ensuring that each NLI item precedes its corresponding VE item to reduce potential priming from the visual evidence.

We focus on SNLI instances that were originally labeled neutral. This choice is motivated by \citet{Kayser_2021_ICCV}, who report that neutral instances are particularly sensitive to replacing textual premises with images. To capture different outcomes of this caption-to-image shift, we use the e-SNLI-VE labels \citep{Kayser_2021_ICCV} to select our 225 instances, balancing them across entailment, neutral, and contradiction in VE. Thus, all instances share a neutral starting point in SNLI, but differ in how they are labeled when the premise is visual. 

Because studying HLV requires multiple judgments, and because our paired comparison requires the same annotators across textual and visual premises, we newly annotate all NLI and VE items. Consequently, the majority labels in \projname{} may differ from the original SNLI and e-SNLI-VE labels.
Figure~\ref{fig:nli_ve_shift} summarizes the resulting label structure, tracing each instance from its original SNLI label to the majority labels from our new NLI and VE annotations. Most instances remain neutral after NLI re-annotation, while 31 shift to entailment. From NLI to VE, 51.1\% of instances change majority label and 48.9\% retain it. Ties denote cases in which no label receives a majority among the five annotators. 

Further details on the annotation procedure, annotator pool, and annotation guidelines are provided in Appendix~\ref{sec:appendix_annotation}, while comparisons with the original e-SNLI-VE labels are reported in Appendix~\ref{appendix:label_shifts_reannotation}.

\section{Experimental Setup}

\subsection{Models}

We evaluate 13 VLMs from five model families on both NLI and VE, including Gemma~3 \citep{gemmateam2025gemma3technicalreport}, InternVL3.5 \citep{wang2025internvl35advancingopensourcemultimodal}, LLaVA-OneVision-1.5 \citep{an2025llavaonevision15fullyopenframework}, Molmo~2 \citep{clark2026molmo2openweightsdata}, and Qwen3-VL \citep{bai2025qwen3vltechnicalreport}, with model sizes ranging from 2B to 27B parameters. 
To contextualize VLM performance on the text-only NLI task, we additionally evaluate Qwen3 \citep{yang2025qwen3technicalreport} as a text-only LLM on the NLI task and report results in Appendix \ref{sec:appendix_text_only_llms}.

\subsection{Prompting and Prediction Extraction}
\label{sec:pred_extraction}

We evaluate models using a zero-shot multiple-choice prompting setup.
To reduce position and option-letter bias \citep{pezeshkpour-hruschka-2024-large, zheng_ICLR2024_mc_robust}, we randomize the mapping between answer letters and labels for each instance. Example prompts are provided in Appendix \ref{sec:appendix_prompting}.
For prediction extraction, let \(Y = \{\mathrm{Ent}, \mathrm{Neu}, \mathrm{Con}\}\) denote the label set. For each instance \(i\), we extract the raw next-token logits for the three answer option letters and map them back to their corresponding labels. This yields label-level scores \(\mathbf{z}_i = [z_i^{\mathrm{Ent}}, z_i^{\mathrm{Neu}}, z_i^{\mathrm{Con}}]\), whose softmax-normalized values define the model judgment distribution \(p_i\) over \(Y\).
The predicted label is the label with the highest probability.

\subsection{Evaluation Metrics}

We summarize the metrics used for the human and model analyses below. Formal definitions and further details are provided in Appendix \ref{sec:appendix_metrics}.

\paragraph{Human Analysis.}
We quantify disagreement using entropy over human label distributions. 
To measure how human judgments change between NLI and VE, we compare the corresponding label distributions using total variation distance (TVD) and Jensen-Shannon distance (JSD), following prior work \citep{baan-etal-2022-stop, nie-etal-2020-learn, chen-etal-2024-seeing, chen-etal-2025-rose}. 
We further analyze human uncertainty using annotator self-reported confidence.

\paragraph{Model Analysis.}
We report both accuracy against the human majority label, and distributional alignment metrics (TVD and JSD).
We further analyze model uncertainty using normalized predictive entropy and conformal prediction with the Least Ambiguous set-valued Classifier (LAC), following \citet{ye_2024_uncertainty}. This method
returns a calibrated set of plausible labels, with larger sets indicating higher uncertainty. In our entailment task, the set size ranges from 1 to 3.
Finally, we compute Spearman correlations between model and human uncertainty.

\begin{figure}[t]
    \centering
    \includegraphics[width=\columnwidth]{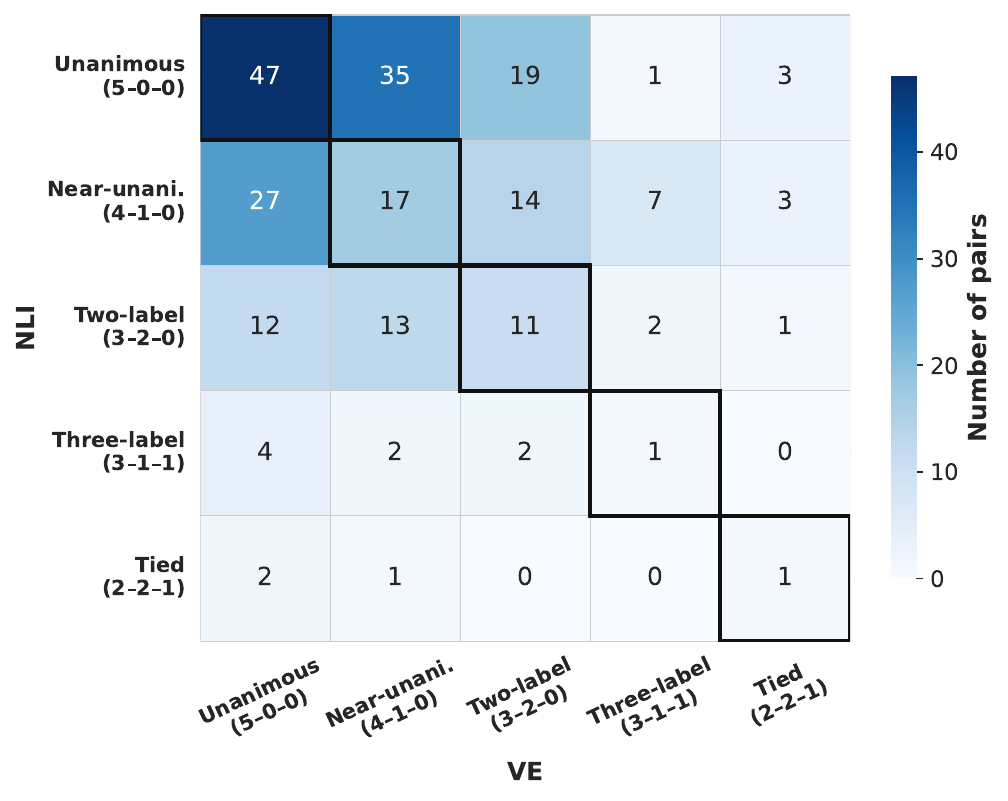}
    \caption{Agreement-pattern transitions NLI $\leftrightarrow$ VE.}
    \label{fig:annotation_pattern_transition}
\end{figure}

\section{Human Label Variation across Textual and Visual Entailment}

\label{sec:hlv_modality_shift}

\subsection{Disagreement Restructuring}
\label{sec:human_disagreement}

To examine how disagreement changes when textual premises are replaced with visual premises, we compare annotation patterns between NLI and VE. We group instances by their five-annotator label distribution: unanimous (5-0-0), near-unanimous (4-1-0), split between two labels (3-2-0), split between three labels (3-1-1), and tied (2-2-1). 
The transition matrix in Figure \ref{fig:annotation_pattern_transition} shows that many instances remain in similar agreement regions across modalities, especially in the high-agreement area. However, the off-diagonal cells show movement in both directions: items to the left of the diagonal become more agreed upon in VE, while items to the right become less agreed upon.

\begin{table}[h!]
\centering
\small
\begin{tabular}{lcc}
\toprule
\textbf{Metric} & \textbf{Mean $\pm$ Std.} & \textbf{95\% CI} \\
\midrule
Entropy (NLI) & $0.47 \pm 0.47$ & [0.41, 0.53] \\
Entropy (VE) & $0.54 \pm 0.49$ & [0.48, 0.60] \\
$\Delta$ entropy (VE--NLI) & $0.07 \pm \text{0.64}$ & [-0.01, 0.15] \\
TVD (VE--NLI) & $0.48 \pm 0.34$ & [0.44, 0.53] \\
JSD (VE--NLI) & $0.52 \pm 0.32$ & [0.48, 0.57] \\
\bottomrule
\end{tabular}
\caption{Human disagreement and NLI--VE distributional shifts with 95\% paired bootstrap confidence intervals. Values are mean $\pm$ standard deviation.}
\label{tab:human_disagreement}
\end{table}

\begin{table*}[th!]
\centering
\small
\setlength{\tabcolsep}{5pt}
\begin{tabular}{@{}llcc@{}}
\toprule
\textbf{What changes between NLI and VE?} &
\textbf{Operational criterion} &
\textbf{$n$} &
\textbf{Disagreement restructuring?} \\
\midrule

$\boldsymbol{D\downarrow}$: Disagreement decreases in VE &
$H_{\mathrm{VE}} < H_{\mathrm{NLI}}$ &
63 &
Yes \\

$\boldsymbol{D\uparrow}$: Disagreement increases in VE &
$H_{\mathrm{VE}} > H_{\mathrm{NLI}}$ &
85 &
Yes \\

$\boldsymbol{D\leftrightarrow}$: Disagreement composition changes &
$H_{\mathrm{NLI}} = H_{\mathrm{VE}} > 0$ and $0 < \mathrm{TVD} < 1$ &
22 &
Yes \\

\midrule

Unanimous label changes &
$H_{\mathrm{NLI}} = H_{\mathrm{VE}} = 0$ and $\mathrm{TVD} = 1$ &
27 &
No \\

Label distribution unchanged &
$H_{\mathrm{NLI}} = H_{\mathrm{VE}}$ and $\mathrm{TVD} = 0$ &
28 &
No \\

\bottomrule
\end{tabular}

\caption{Operationalization of disagreement restructuring across paired NLI--VE instances ($n=225$). We consider disagreement restructured when its amount changes, or when the amount remains the same but the label distributions differ. $H$ denotes human label entropy, while $n$ indicates item count. TVD is total variation distance.}
\label{tab:disagreement_restructuring}
\end{table*}

Table \ref{tab:human_disagreement} summarizes the changes using human label entropy and cross-modal distributional distances.
At the aggregate level, we find no clear evidence that visual premises systematically increase or decrease disagreement, as the difference in mean human label entropy between NLI and VE is small and not statistically distinguishable from zero.
Importantly, at the instance level, TVD and JSD scores indicate substantial changes in how judgments are distributed across labels.\footnote{Additional stability checks are reported in Appendix \ref{appendix:stability_checks}.}
Not every distributional change, however, constitutes a change in disagreement. An instance may, for example, switch unanimously from one label to another while remaining fully agreed upon in both modalities. We therefore define disagreement as \textit{restructured}, see Table \ref{tab:disagreement_restructuring}, when either (i) its amount changes across modalities, measured by a change in item-level entropy, or (ii) disagreement is present in both modalities and its label composition changes despite an unchanged entropy level.
Under this operationalization, we observe that 170 of 225 instances (75.6\%) exhibit disagreement restructuring. For 12\%, the label distribution remains unchanged across modalities, while for the other 12\% labels flips unanimously.

\begin{figure}[t]
    \centering
    \includegraphics[width=\linewidth]{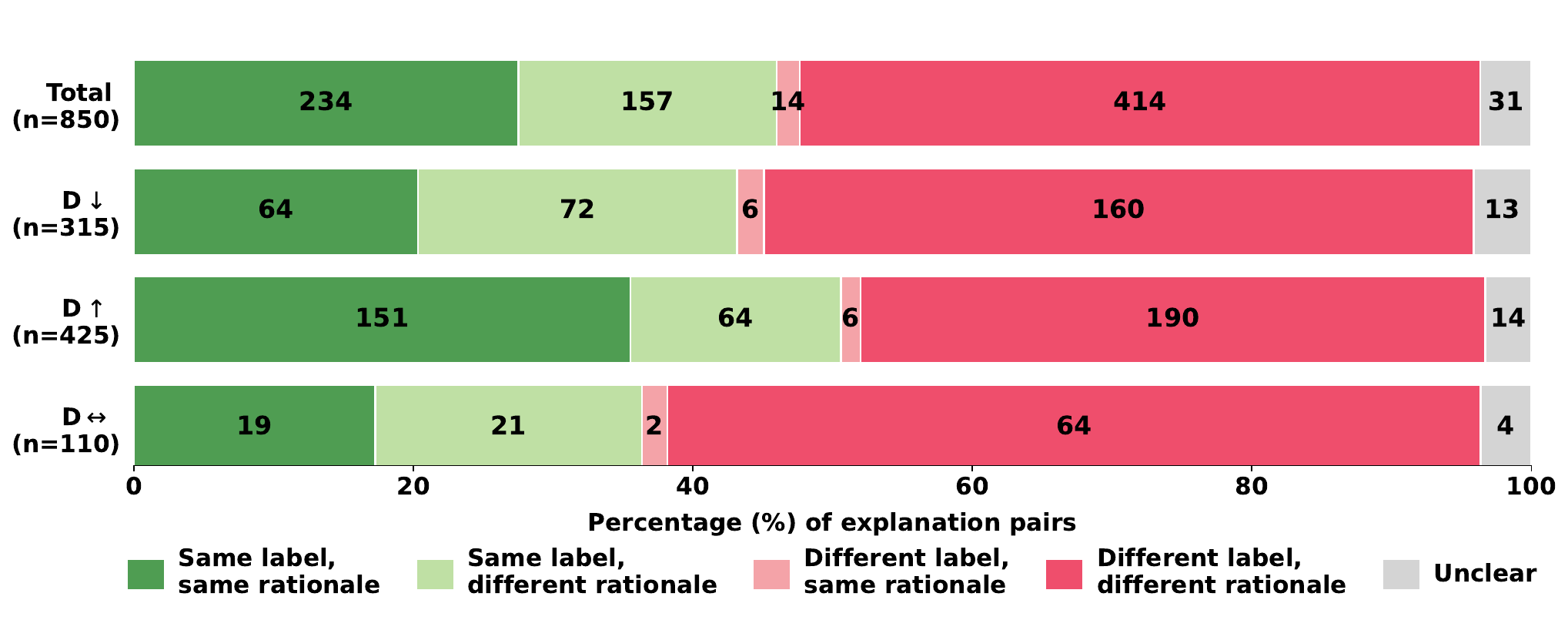}
    \caption{Explanation analysis for instances exhibiting disagreement restructuring. Numbers indicate explanation-pair counts. $\boldsymbol{D\downarrow}$: Disagreement decreases in VE; $\boldsymbol{D\uparrow}$: Disagreement increases in VE; $\boldsymbol{D\leftrightarrow}$: Disagreement composition changes.}
    \label{fig:explanation_analysis}
\end{figure}

\begin{figure*}[th!]
    \centering
    \includegraphics[width=\textwidth]{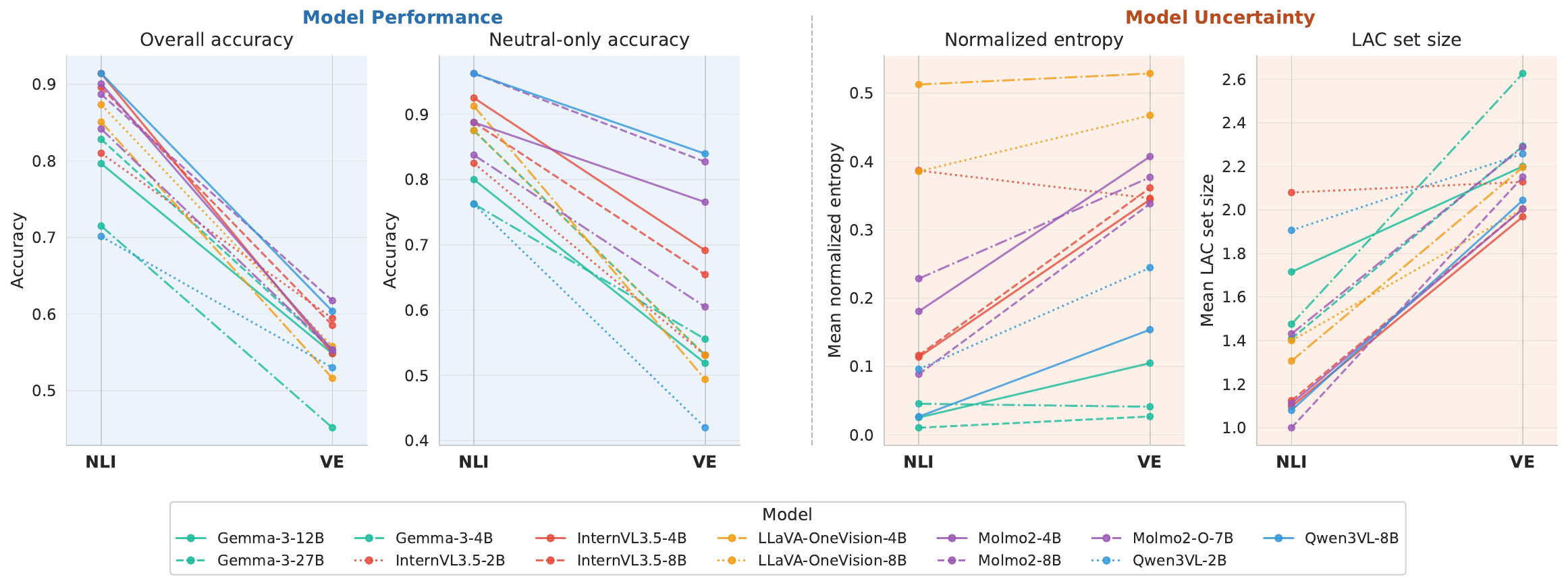}
    \caption{
    Model performance and uncertainty across NLI and VE. The left half shows accuracy with respect to human majority labels across all instances and on the neutral-only paired instances. The right half shows model uncertainty, measured by normalized predictive entropy and LAC conformal prediction set size, with larger LAC sets indicating higher uncertainty. Accuracy decreases from NLI to VE, while uncertainty generally increases.
    }
    \label{fig:model_perf_uncertainty}
\end{figure*}

\subsection{HLV through the Lens of Explanations}

Label distributions capture changes in assigned labels, but not the reason behind them. 
Because in \textsc{VariNLI-VE} the same annotators provide explanations, our dataset allows us to examine whether the same label is supported by a similar reason across the textual (NLI) and visual premises (VE). For the 170 instances exhibiting disagreement restructuring, we manually classify all 850 paired NLI–VE explanations (170 instances $\times$ 5 annotators) by whether the label and rationale remain the same or change. One author classified all 850 pairs, while a second independently classified 101 of these pairs to assess reliability (89.1\% agreement; Cohen's $\kappa = 0.74$).
Pairs were marked unclear when explanations were too brief or generic to determine a rationale change.

Figure~\ref{fig:explanation_analysis} summarizes the resulting patterns. As expected, label changes are usually accompanied by changes in rationale, accounting for 414 pairs (in red), with only a few exceptions.
Another 234 keep both the same label and a similar rationale. More notably, 157 pairs (light green) retain the same label but change rationale. These label-preserving rationale shifts show that changes in the available evidence can alter an annotator's rationale without changing their label. This echoes within-label variation observed in text-based NLI \citep{jiang-etal-2023-ecologically, hong-etal-2025-litex}, but here it occurs within the same annotator across modalities.
This shows the importance of explanation, as they reveal variation that remains invisible at the label level. Examples of all patterns are shown in Appendix~\ref{sec:appendix_explanation_examples}.

\subsection{Confidence and Disagreement}

While entropy captures between-annotator disagreement, self-reported confidence captures annotators' subjective certainty in their own judgments.
This raises the question whether disagreement is accompanied by lower confidence, and how confidence is affected by visual grounding.
Table~\ref{tab:confidence_by_pattern} shows that human confidence (ranging from 1-6) is generally high in both NLI and VE and similar across the two settings within the same agreement pattern.
Replacing the textual premise with an image therefore does not appear to make annotators substantially more confident.
Instead, confidence varies more clearly with the degree of agreement.
Unanimously labeled instances receive the highest confidence scores in both settings, while divisive patterns obtain lower scores.
At the same time, confidence remains relatively high even for split cases, suggesting that HLV can reflect confidently held competing interpretations.
We therefore treat confidence and disagreement as related but complementary signals of human uncertainty.

\begin{table}[t!]
\centering
\small
\begin{tabular}{lccc}
\toprule
\textbf{Pattern} & \textbf{NLI} & \textbf{VE} & \textbf{$n$ (NLI/VE)} \\
\midrule
5-0-0 & $5.39 \pm 0.49$ & $5.41 \pm 0.54$ & 105/92 \\
4-1-0 & $4.94 \pm 0.60$ & $5.00 \pm 0.52$ & 68/68 \\
3-2-0 & $4.67 \pm 0.58$ & $4.76 \pm 0.44$ & 39/46 \\
3-1-1 & $4.82 \pm 0.41$ & $4.69 \pm 0.40$ & 9/11 \\
2-2-1 & $4.75 \pm 0.77$ & $4.62 \pm 0.68$ & 4/8 \\
\bottomrule
\end{tabular}
\caption{Mean confidence by annotation pattern in NLI and VE. Values are mean $\pm$ standard deviation on a six-point confidence scale, while $n$ indicates item count.}
\label{tab:confidence_by_pattern}
\end{table}

\section{Model Behavior across Textual and Visual Entailment}

\subsection{Performance and Prediction Patterns} \label{sec:performance_pred_patterns}

\begin{table}[t]
\centering
\small
\setlength{\tabcolsep}{3pt}
\begin{tabular}{l c | c | c c c}
\toprule
\multirow{2}{*}{\textbf{Model}} & \multirow{2}{*}{\textbf{NLI}} & \multicolumn{1}{c|}{\textbf{VE}} & \multicolumn{3}{c}{\textbf{VE (by label)}} \\
\cmidrule(lr){4-6}
 &  & \textbf{Overall} & \textbf{Ent} & \textbf{Neu} & \textbf{Con} \\
\midrule

Gemma-3-4B  & 71.5 & \ws{26}{45.2} & \ws{0}{87.7} & \ws{16}{55.6} & \ws{68}{3.8} \\
Gemma-3-12B & 79.6 & \ws{25}{54.8} & \ws{0}{87.7} & \ws{28}{51.9} & \ws{45}{\underline{34.2}} \\
Gemma-3-27B & 82.8 & \ws{28}{55.3} & \ws{0}{87.7} & \ws{30}{53.1} & \ws{49}{\underline{34.2}} \\

InternVL3.5-2B & 81.0 & \ws{22}{59.4} & \ws{0}{93.0} & \ws{28}{53.1} & \ws{39}{\textbf{41.8}} \\
InternVL3.5-4B & \textbf{91.4} & \ws{37}{54.8} & \ws{2}{89.5} & \ws{22}{69.1} & \ws{76}{15.2} \\
InternVL3.5-8B & 89.6 & \ws{31}{58.5} & \ws{0}{94.7} & \ws{24}{65.4} & \ws{64}{25.3} \\

LLaVA-OneVision-4B & 85.1 & \ws{34}{51.6} & \ws{0}{\textbf{100.0}} & \ws{36}{49.4} & \ws{66}{19.0} \\
LLaVA-OneVision-8B & 87.3 & \ws{32}{55.8} & \ws{0}{\underline{98.2}} & \ws{34}{53.1} & \ws{60}{27.8} \\

Molmo2-4B  & \underline{90.0} & \ws{35}{55.3} & \ws{6}{84.2} & \ws{13}{76.5} & \ws{77}{12.7} \\
Molmo2-8B  & 88.7 & \ws{27}{\textbf{61.8}} & \ws{1}{87.7} & \ws{6}{\underline{82.7}} & \ws{67}{21.5} \\
Molmo2-O-7B & 84.2 & \ws{29}{55.3} & \ws{0}{93.0} & \ws{24}{60.5} & \ws{61}{22.8} \\

Qwen3VL-2B & 70.1 & \ws{17}{53.0} & \ws{0}{96.5} & \ws{28}{42.0} & \ws{37}{32.9} \\
Qwen3VL-8B & \textbf{91.4} & \ws{31}{\underline{60.4}} & \ws{11}{80.7} & \ws{7}{\textbf{84.0}} & \ws{70}{21.5} \\

\bottomrule
\end{tabular}
\caption{
Model accuracies across modalities and by VE majority label. Cells are shaded according to decrease in accuracy wrt NLI; darker cells indicate larger drops.
}
\label{tab:ve_labelwise}
\end{table}

\paragraph{Accuracy consistently drops from NLI to VE.}
We first evaluate each model against the majority label from our new NLI and VE annotations. 
Figure~\ref{fig:model_perf_uncertainty} (left panel) shows consistently higher accuracy for NLI than VE across model families. The strongest models reach high accuracy on NLI, with InternVL3.5-4B and Qwen3VL-8B both achieving above 90\% accuracy against the human majority label. In contrast, Molmo2-8B only reaches 61.8\% accuracy as the best performing VE model.
To better understand this large gap, we further break accuracy down by VE majority label. Table~\ref{tab:ve_labelwise} shows that models perform well on entailment, with several models scoring close or above 90\%. However, performance is lower on neutral instances and drops severely for contradiction. This suggests that current VLMs are effective at identifying visual evidence that supports a hypothesis, but less reliable at detecting visual evidence that contradicts it.
This pattern is further supported by the aggregate model prediction distribution (App.\ Figure
\ref{fig:appendix_model_label_dist}), showing that models underpredict contradiction and instead assign a larger share to entailment or neutral.

\paragraph{The modality gap remains for neutral cases.}
Since our NLI subset is neutral-focused while the corresponding VE annotations have a more balanced distribution, the NLI-VE accuracy drop may partly reflect this label shift. To assess this, we examine the subset of paired instances whose VE majority label is neutral, and compare model accuracy with the corresponding NLI instances. As shown in the second plot in Figure \ref{fig:model_perf_uncertainty}, even on this controlled subset, all VLMs still perform worse in VE than in NLI, with drops ranging from 12.2\% for Molmo2-4B to 41.9\% for LLaVA-OneVision-4B. While this indicates that the modality gap cannot be explained by label shift alone, other factors, such as potential textual shortcuts in NLI or uneven reliance on textual versus visual evidence, may also contribute. Disentangling these factors is beyond the scope of the present analysis but remains an interesting direction for future work.
Full per-model results are reported in Appendix~\ref{sec:appendix_neutral_accuracies}, while Appendix~\ref{appendix:cis} reports the 95\% confidence intervals.

\begin{figure}[t]
    \centering
    \includegraphics[width=0.9\linewidth]{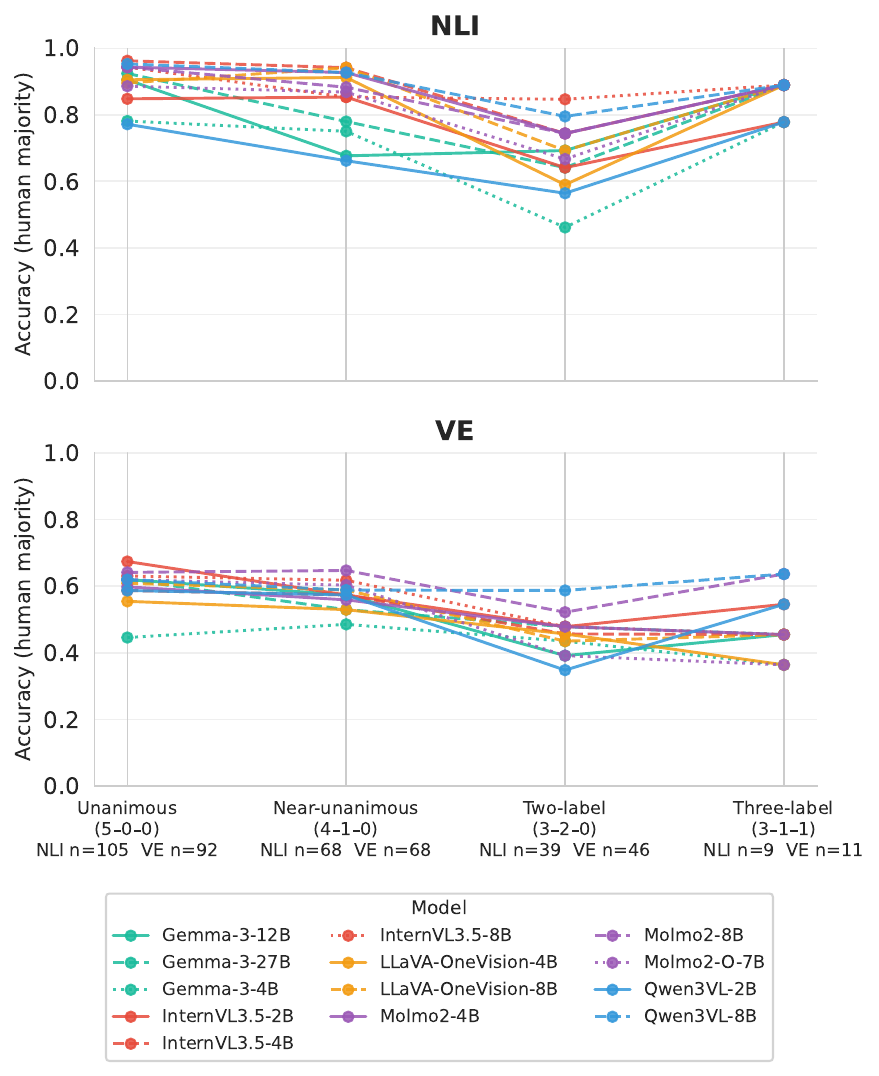}
    \caption{Per-model accuracy by annotator agreement.}
    \label{fig:accuracy_by_agreement}
\end{figure}

\paragraph{Accuracy decreases with disagreement.}
Figure~\ref{fig:accuracy_by_agreement} shows model accuracy by human annotator agreement pattern. Across modalities, accuracy is highest on unanimous or near-unanimous instances and lower on instances with greater disagreement, especially in VE.
Additionally, VE accuracy remains lower than NLI accuracy across agreement patterns, indicating that the modality gap is not limited to highly ambiguous instances.

\subsection{Alignment with Human Judgments}

So far, we have considered model behavior through aggregate prediction patterns, but these do not show whether model predictive distributions capture instance-level variation in human judgments. 
We therefore compare each model's predicted label distributions to human judgment distributions to inspect whether models reflect HLV rather than only the majority vote.
Following prior work \citep{baan-etal-2022-stop, nie-etal-2020-learn, chen-etal-2025-rose, chen-etal-2024-seeing}, we use TVD and JSD and report results in Table \ref{tab:human_model_alignment}. 
Across all model families, model predictive distributions align more closely with human label distributions in NLI than in VE, with higher TVD and JSD in VE for every model. 
The best-aligned models are InternVL3.5-4B in NLI and InternVL3.5-8B in VE for both metrics.
However, even the best VE alignment scores remain considerably lower compared to the best NLI scores.
As with accuracy, this gap persists on the neutral-only subset. Even when we restrict the comparison to paired instances whose VE majority label is neutral, all VLMs have higher TVD and JSD in VE than on the corresponding NLI examples.
A label-wise VE breakdown in Table \ref{tab:appendix_ve_labelwise_alignment} in the Appendix shows that divergence is highest for contradiction cases, consistent with the low accuracy and the underprediction of this label observed above.

\begin{table*}[t]
\centering
\small
\setlength{\tabcolsep}{2pt}
\begin{tabular*}{\textwidth}{l @{\extracolsep{\fill}} cc cc cc cc}
\toprule
\multirow{2}{*}{\textbf{Model}}
& \multicolumn{4}{c}{\textbf{All instances}}
& \multicolumn{4}{c}{\textbf{Neutral-only instances}} \\
\cmidrule(lr){2-5} \cmidrule(lr){6-9}
& \textbf{TVD NLI $\downarrow$} & \textbf{TVD VE $\downarrow$}
& \textbf{JSD NLI $\downarrow$} & \textbf{JSD VE $\downarrow$}
& \textbf{TVD NLI $\downarrow$} & \textbf{TVD VE $\downarrow$}
& \textbf{JSD NLI $\downarrow$} & \textbf{JSD VE $\downarrow$} \\
\midrule

Gemma-3-4B         
& 0.325 & \dcell{24}{0.561} 
& 0.367 & \dcell{23}{0.593}
& 0.278 & \dcell{22}{0.497} 
& 0.319 & \dcell{22}{0.541} \\

Gemma-3-12B        
& 0.268 & \dcell{19}{0.460} 
& 0.309 & \dcell{19}{0.500}
& 0.245 & \dcell{24}{0.481} 
& 0.286 & \dcell{24}{0.528} \\

Gemma-3-27B        
& 0.245 & \dcell{23}{0.477} 
& 0.289 & \dcell{22}{0.508}
& 0.204 & \dcell{29}{0.492} 
& 0.248 & \dcell{28}{0.531} \\

InternVL3.5-2B     
& 0.253 & \dcell{14}{\underline{0.392}} 
& 0.322 & \dcell{11}{\underline{0.434}}
& 0.255 & \dcell{20}{0.454} 
& 0.331 & \dcell{17}{0.500} \\

InternVL3.5-4B     
& \textbf{0.185} & \dcell{22}{0.407} 
& \textbf{0.244} & \dcell{20}{0.443}
& 0.148 & \dcell{18}{0.329} 
& \underline{0.211} & \dcell{17}{0.382} \\

InternVL3.5-8B     
& 0.196 & \dcell{19}{\textbf{0.388}} 
& 0.249 & \dcell{18}{\textbf{0.426}}
& 0.177 & \dcell{16}{0.341} 
& 0.233 & \dcell{16}{0.392} \\

LLaVA-OneVision-4B 
& 0.224 & \dcell{21}{0.430} 
& 0.301 & \dcell{17}{0.470}
& 0.226 & \dcell{21}{0.432} 
& 0.313 & \dcell{16}{0.474} \\

LLaVA-OneVision-8B 
& 0.204 & \dcell{20}{0.399} 
& 0.278 & \dcell{16}{0.440}
& 0.203 & \dcell{19}{0.398} 
& 0.282 & \dcell{16}{0.439} \\

Molmo2-4B          
& \underline{0.191} & \dcell{22}{0.407} 
& 0.253 & \dcell{19}{0.444}
& 0.188 & \dcell{10}{\underline{0.290}} 
& 0.257 & \dcell{9}{\textbf{0.346}} \\

Molmo2-8B          
& 0.203 & \dcell{20}{0.401} 
& 0.261 & \dcell{18}{0.446}
& \underline{0.146} & \dcell{14}{\textbf{0.287}} 
& 0.213 & \dcell{14}{\underline{0.349}} \\

Molmo2-O-7B        
& 0.224 & \dcell{19}{0.410} 
& 0.287 & \dcell{17}{0.453}
& 0.219 & \dcell{17}{0.388} 
& 0.278 & \dcell{16}{0.441} \\

Qwen3VL-2B         
& 0.337 & \dcell{11}{0.447} 
& 0.377 & \dcell{10}{0.481}
& 0.296 & \dcell{22}{0.517} 
& 0.343 & \dcell{21}{0.551} \\

Qwen3VL-8B         
& 0.197 & \dcell{22}{0.418} 
& \underline{0.248} & \dcell{21}{0.454}
& \textbf{0.144} & \dcell{15}{0.291} 
& \textbf{0.197} & \dcell{16}{0.352} \\

\bottomrule
\end{tabular*}
\caption{
Alignment between model predictive distributions and human judgment distributions. Lower TVD and JSD indicate closer alignment. VE columns are shaded according to the increase over the corresponding NLI value within each block, with darker cells indicating a larger gap.
}
\label{tab:human_model_alignment}
\end{table*}

\subsection{Model Uncertainty}

We first analyze model uncertainty across NLI and VE using normalized predictive entropy and LAC set size. As shown in Figure~\ref{fig:model_perf_uncertainty} (right panel), uncertainty is generally higher in VE: predictive entropy is higher for 11/13 models, with 95\% CIs excluding zero for 10 models. For LAC set size, all 13 show higher uncertainty, with CIs excluding zero for 12 models.
Full CIs are reported in Appendix~\ref{appendix:cis}, and detailed model uncertainty scores in Appendix~\ref{sec:appendix_full_uncertainty}.

\begin{figure}[t!]
    \centering
    \includegraphics[width=0.8\linewidth]{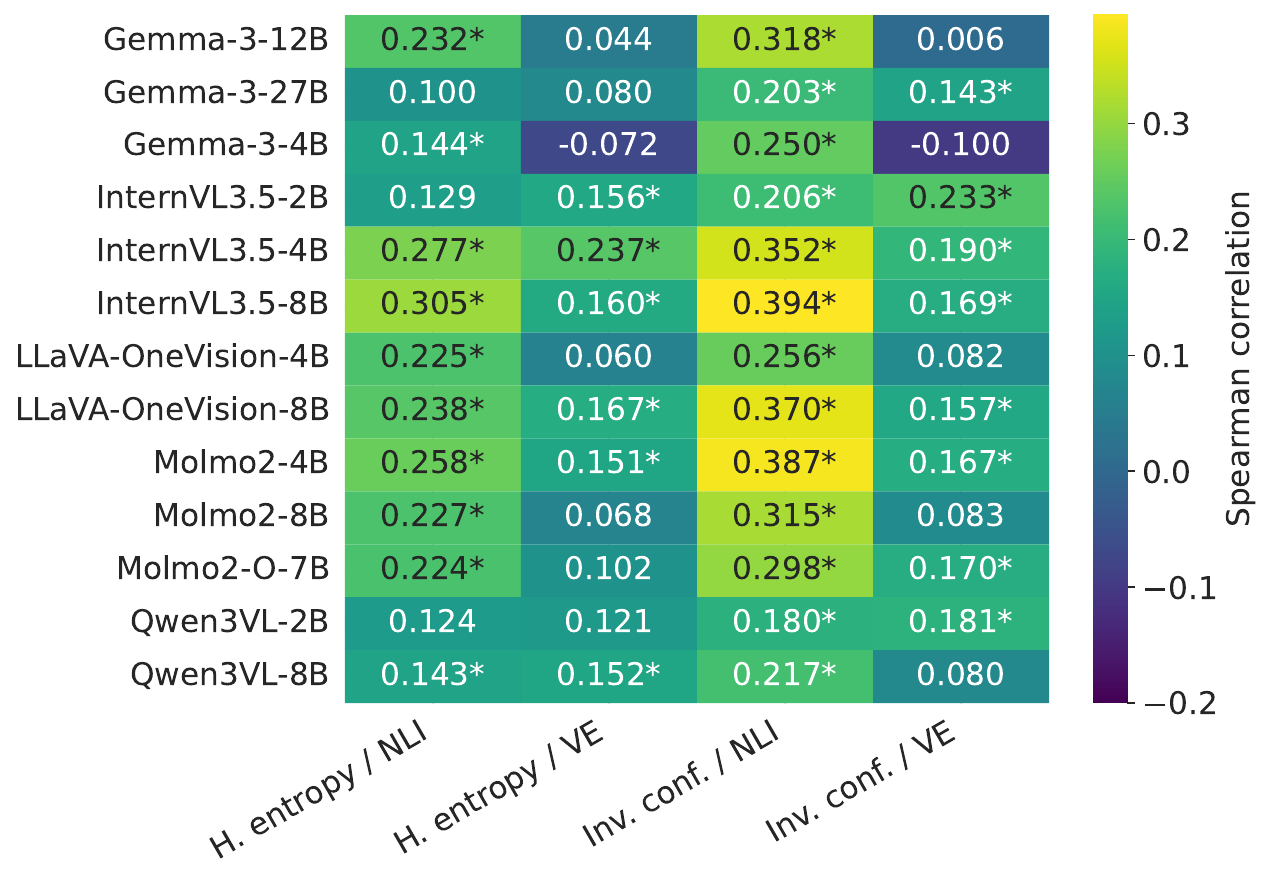}
    \caption{
    Spearman's
    $\rho$ between model predictive entropy and human uncertainty measures (human label entropy and inverse mean confidence). Asterisks indicate statistically significant correlations (p < 0.05).}
    \label{fig:uncertainty_alignment_entropy}
\end{figure}

To assess whether models tend to be more uncertain on instances that humans also find more uncertain, we compute Spearman correlations with human label entropy and inverse confidence.
Figure~\ref{fig:uncertainty_alignment_entropy} shows that predictive entropy correlates only weakly with human uncertainty, with median $\rho = 0.235$ in NLI and $0.147$ in VE. The associations are statistically significant for 88\% of NLI correlations, but only 54\% of VE correlations. LAC set size correlations, reported in Figure~\ref{fig:appendix_uncertainty_alignment_lac} in the Appendix, are similarly weak (71\% significant in NLI; 50\% in VE). Across both measures, model uncertainty only weakly tracks human uncertainty, with significant associations occurring less consistently in VE.

\section{Discussion}

We asked how HLV changes when a premise in NLI is replaced by an image, and whether VLMs reflect the resulting human judgment patterns. Our findings suggest that visual evidence does not simply resolve ambiguity. It can clarify information left underspecified in the text, but also introduce additional aspects of the scene that annotators interpret differently. Confidence provides a complementary view, as annotators can remain confident even in contested cases. Explanations reveal another layer of variation, showing that an annotator’s rationale can change even when the label remains the same.

These findings matter for HLV research, as a majority label cannot distinguish broad agreement from contested cases, while labels alone cannot capture changes in the rationale behind a judgment. We consider label distributions, confidence, and explanations across paired NLI and VE instances, allowing HLV to be studied at several levels rather than reducing it to disagreement over the final label.
This is also important for model evaluation. VLMs align less closely with human judgment distributions in VE, while model uncertainty only weakly tracks human uncertainty. An evaluation based only on majority labels can hide whether a model captures the range of human judgments or merely predicts the dominant label.

More broadly, these findings motivate datasets that preserve meaningful variation, disagreement-aware evaluation, and systematic analyses of why judgments change from NLI to VE. Such work could also help identify cases where VLM predictions reflect or amplify social biases, contributing to more transparent and reliable VLMs.

\section{Conclusion}

We study HLV in a paired NLI–VE setting, where the same hypothesis is judged against either a textual premise or an image. We introduce VariNLI-VE, with 2,250 labels, explanations, and confidence scores from the same annotators for originally neutral SNLI instances. Contrary to the expectation that visual evidence would reduce disagreement, we find no clear overall decrease: disagreement decreases for some instances while increasing for others, and explanations reveal rationale changes even when labels remain unchanged. VLMs show lower accuracy, weaker alignment with human judgments, and generally higher uncertainty in VE. We hope our work encourages further research into why HLV changes across textual and visual evidence, and how these changes relate to model behavior.

\section*{Limitations}

One limitation of our work is the scale of our dataset due to resource constraints and the preference to collect high-quality annotations without having to resort to crowdsourcing. 
While we collect 2,250 annotations, these cover 225 underlying instances with five judgments each for NLI and VE.
A larger dataset with more annotators per instance would indeed provide a more stable estimate of HLV.
Still, we believe our paired same-annotator design, with labels, confidence, and explanations for both NLI and VE, provides a useful resource for studying HLV across textual and visual evidence.

Another limitation is that we focus on originally neutral NLI instances, since these are the most sensitive to the introduction of visual evidence.
This allows us to target cases where a caption-to-image shift in the premise is likely to matter, but it remains open whether similar HLV patterns occur for originally entailment or contradiction instances.

Furthermore, our analysis focuses mainly on aggregate judgment patterns and does not analyze annotator-level variation.
Annotators may differ systematically in how they interpret labels or how they use confidence scores.
While the released dataset provides the opportunity to investigate whether some annotators are more sensitive to visual evidence than others, we leave this question to future work.

Among our pool of 21 annotators, three were authors of this paper, although they never constituted a majority for any instance. We assess this potential source of bias through an author-exclusion analysis and find the main findings persisting even after excluding author-annotators 
(cf. Appendix~\ref{appendix:stability_checks}). 
Further, as described in Appendix~\ref{sec:appendix_annotator_metadata}, most annotators were non-native English speakers, although they generally acquired English early, reported high proficiency, and were allowed to look up unfamiliar terms. Future work could assess whether similar patterns hold with an annotator pool of native speakers.

Our explanation analysis provides an initial account of whether annotator rationales change across textual and visual premises, but does not aim to provide an exhaustive taxonomy of the sources and reasons of disagreement. A more in-depth analysis, as well as a comparison of human and model explanations, remain important future directions 
enabled by \projname{}.

\section*{Ethical Considerations}

We study HLV in a paired NLI and VE setup, with the goal of understanding how human judgments and model predictions change when a textual premise is replaced by an image.
While we do not foresee direct harmful implications specific to our dataset or analyses, broader ethical considerations remain.
Vision-language models may reproduce or amplify social biases present in their training data, which may affect both their predictions and uncertainty estimates.
Despite this risk, we believe that explicitly studying, preserving, and modeling human label variation can have a positive downstream impact by informing the development of more transparent and less biased models.

We acknowledge the use of AI tools to support parts of the code development and writing refinement. All AI-generated outputs were carefully reviewed, and we confirm that we take full responsibility for the final content and results.

\section*{Acknowledgments}

We thank the members of the MaiNLP lab for their insightful feedback and discussions, particularly Diego Frassinelli, Michael Hedderich, Mahdi Dhaini, Shijia Zhou and Andreas Säuberli. We are also grateful to the participants in the NLP course at the Free University of Bozen-Bolzano for their valuable contribution to the annotation of our dataset. Furthermore, we thank the anonymous reviewers for their constructive feedback, which helped improve this paper. We acknowledge funding through the ERC Consolidator Grant DIALECT 101043235, and the German Research Foundation (DFG) project VariVLM (586887264).

\bibliography{custom}

\appendix

\section{Annotation Details}
\label{sec:appendix_annotation}

\subsection{Annotation Procedure}

Before starting their annotation session, annotators were given a short oral briefing on the task, the corresponding labels \emph{entailment}, \emph{contradiction}, and \emph{neutral}, as well as the annotation process in general. To ensure their consent and to be able to collect their metadata, they were asked to fill out an online form, using previously assigned annotator codes as identifiers. These codes facilitated individual annotator removal in case of faulty annotations or annotators not following instructions, ensuring the quality of annotations.

For annotators previously unfamiliar with annotation or entailment tasks, we offered the opportunity to explore the tool and task via a dedicated \emph{playground} interface. This interface was designed to closely match the interface of the real annotation task. The instances presented in it were carefully selected to not overlap with the actual to-be-annotated instances, yet show a broad overview of different types of instances and give annotators an impression of what would be expected in the task.

In the playground interface, as well as in the real task that followed shortly after, written annotation guidelines were shown as soon as annotators opened the interface (for more details see~\ref{appendix:annotation_guidelines}) and were available to them throughout the whole duration of the task. Moreover, annotators were encouraged to raise any questions they may have and were explicitly allowed to look up any unknown terms they might encounter.

Each annotator was allocated a subset of our data, ensuring that each instance was labeled by exactly five annotators. Moreover, the subsets ensured that one annotator worked on both variants of a given NLI-VE instance, such that potential label shifts from the textual to the visual modality would remain as comparable as possible.
To this end, the order of instances in each subset was pseudo-randomized. While the appearance of items was generally random, some restrictions were put in place: 1) NLI instances were always shown before their corresponding VE versions, as we assumed seeing the multi-modal VE variants prior to the textual ones might prime or bias annotators to favor interpretations more closely aligned with the VE ones. 2) Paired versions were not shown consecutively, again to reduce any potential priming or biasing effects of interpretations trickling from one variant to the other.

The annotation task consisted of three parts: 1) label, 2) explanation, and 3) confidence scores (see also Figures~\ref{fig:nli_interface} and~\ref{fig:ve_interface}). For 1) label, annotators were asked to choose a label out of entailment, contradiction, and neutral, which they then justified in 2) by providing a short rationale on why they chose said label. As we expected some genuinely ambiguous or hard-to-judge instances, especially in NLI, we additionally included 3) confidence scores. Here, annotators indicated on a scale from 1-6 how certain or uncertain they were about having chosen their provided label.

\subsection{Annotator Metadata} \label{sec:appendix_annotator_metadata}

\begin{figure}[t]
    \centering
    \includegraphics[width=\columnwidth]{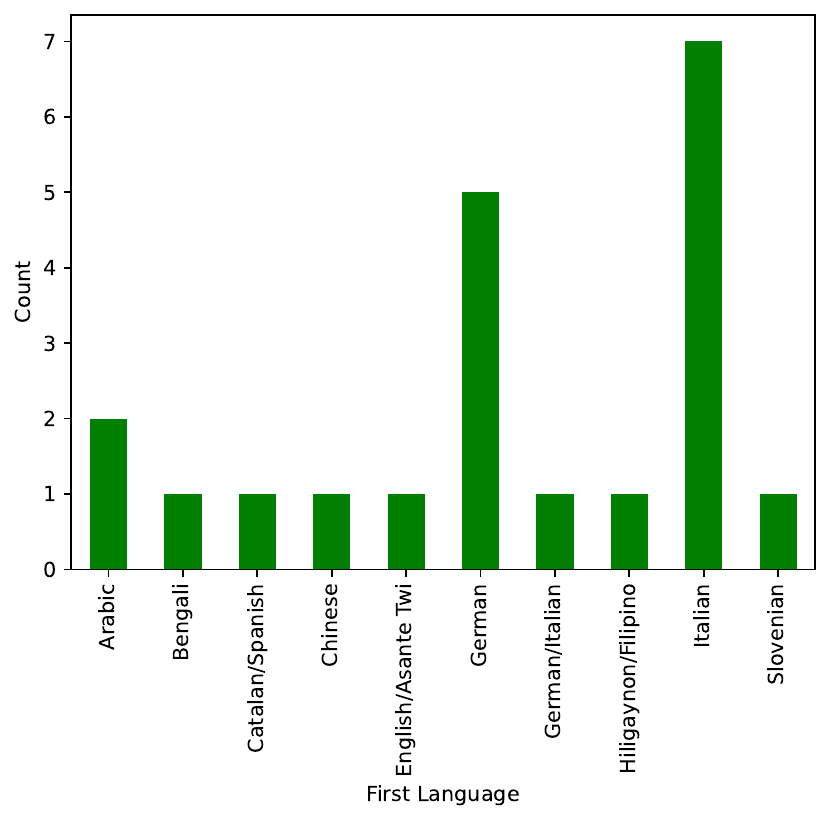}
    \caption{Distribution of the first languages of the annotators who participated in this study.}
    \label{fig:annotator_L1s}
\end{figure}

\begin{table}[t]
\centering
\small
\setlength{\tabcolsep}{4pt}
\begin{tabular}{l ccc cc}
\toprule
\textbf{Metric} & \textbf{Mean} & \textbf{Std.} & \textbf{Range} \\
\midrule
AoA & 7.71 & 2.65 & 3 - 14 \\
Reading Proficiency & 5.71 & 0.46 & 5 - 6 \\
Writing Proficiency & 5.24 & 0.62 & 4 - 6 \\
\bottomrule
\end{tabular}
\caption{Annotator metadata: Age of Acquisition (AoA) of English, as well as self-assessed English reading and writing proficiency of the annotators used in this study. We report mean, standard deviation (std) and range. Proficiency was measured on a 6-point scale with 1 indicating minimal and 6 high proficiency.}
\label{tab:annotator_metadata}
\end{table}

Annotator metadata was collected via the online forms between the oral introduction and starting the annotation.
In the form, annotators first provided written consent for participating in the study and their annotations to be used within the scope of research.
In total, 22 annotators were recruited in this paper, out of which 14 were 1st year M.Sc. computer science students, who participated as part of a university course, and the rest were in-house annotators consisting of three authors of this paper and 5 hired PhD students and post-docs.
After the removal of one annotator for poor performance, the remaining set consisted of 21 annotators in total, resulting in 2,250 annotations overall.

Figure~\ref{fig:annotator_L1s} shows the distribution of native languages in our annotator sample. 7 annotators were native Italians, 5 native Germans and one was German-Italian. The remaining 8 annotators were from other non-English speaking backgrounds such as Arabic, Bengali, or Slovenian, or reported a mixed upbringing, such as English-Asante Twi.

Although most of the annotators employed in this study are not English natives, most reported learning English very early (mean age of acquisition (AoA): $7.71$, std: $2.65$; see Table~\ref{tab:annotator_metadata}). Additionally, their self-reported English proficiency scores, assessed via a 6-point scale across reading and writing skills, indicate a generally high proficiency among all annotators. As shown in Table~\ref{tab:annotator_metadata}, mean reading proficiency sits at $5.71$ (std: $0.46$, range: $5-6$), while writing proficiency presents only marginally lower at $5.24$ (std: $0.62$, range: $4-6$). In addition to self-reported proficiency measures generally being an appropriate approximation of actual language skills \citep{bachman1989construct, li2021meta, wilson1999validity}, annotators were also explicitly allowed to look up any terms they might not know, altogether suggesting sufficient language proficiency and resources to provide adequate annotations.

\begin{figure}[t]
\small
\centering
\begin{tcolorbox}[width=\columnwidth]

\textbf{The task}

For each sample, you are asked to provide: \\
- \textbf{label} (entailment, contradiction, or neutral) \\
- short \textbf{explanation} for the label (1 short sentence) \\
- \textbf{confidence} score (1-6)

\medskip

\textbf{Label}

You will be asked to select a label based on the relationship between a \textbf{premise} and a \textbf{hypothesis}. The possible relationships between a premise and a hypothesis are: \\
- \textbf{Entailment}: The hypothesis is definitely true given the premise. \\
- \textbf{Contradiction}: The hypothesis is definitely false given the premise. \\
- \textbf{Neutral}: The premise does not provide enough information to determine whether the hypothesis is definitely true or false.

\medskip

\textbf{Explanation}

After choosing your label, we will ask you to give a short explanation for why you chose your label. \\
One sentence is enough.

\medskip

\textbf{Confidence score}

Then, please assign a score indicating how confident you feel about the selected label: \\
- 1 = not confident at all \\
- 6 = very confident

\medskip

If you are done with an example, submit your annotations by clicking on <submit>. Please do not skip examples as all examples should be annotated.

\medskip

Note: You can always come back to these instructions by clicking on "instructions" in the top right corner of your screen.

\end{tcolorbox}
\caption{Annotation guidelines that were presented to annotators before starting their task and available to them throughout the session.}
\label{fig:annotation_guidelines}
\end{figure}

\subsection{Annotation Guidelines} \label{appendix:annotation_guidelines}
Figure~\ref{fig:annotation_guidelines} depicts the annotation guidelines given to annotators prior to starting their task.
These guidelines explain the three labels entailment, contradiction, and neutral and give instructions on providing explanations and confidence scores.
Further, annotators were provided with information on how to save their annotations, as well as how to get back to the guidelines if they ever needed to.

\subsection{Annotation Interface}
\begin{figure}[t]
    \centering
    \includegraphics[width=\columnwidth]{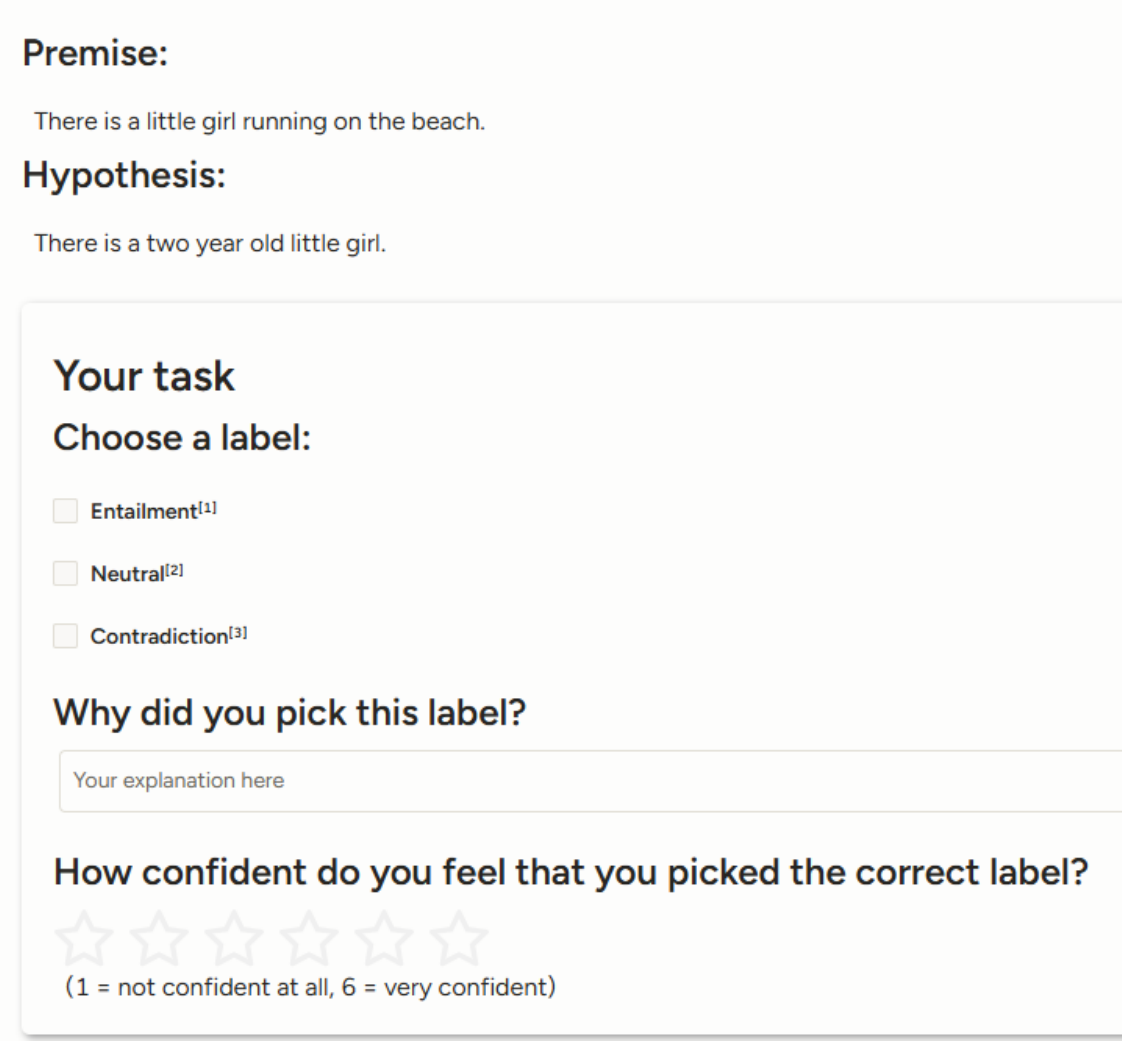}
    \caption{Annotation interface presenting a textual (NLI) instance.}
    \label{fig:nli_interface}
\end{figure}

\begin{figure}[th!]
    \centering
    \includegraphics[width=\columnwidth]{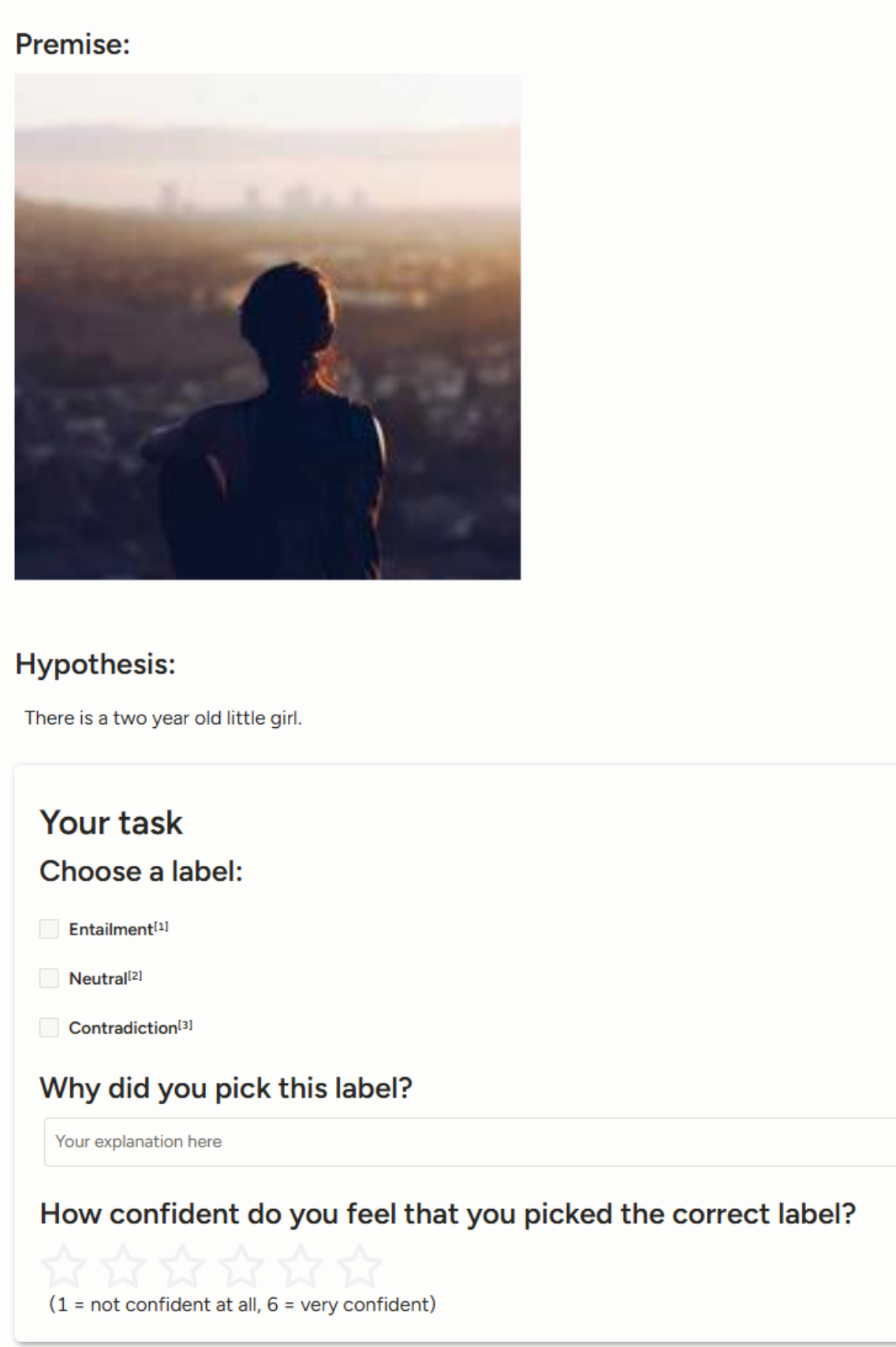}
    \caption{Annotation interface presenting a visual (VE) instance.}
    \label{fig:ve_interface}
\end{figure}

We used LabelStudio~\citep{LabelStudio} for its simple integration of both textual and visual content, while providing remote accessibility to annotators via server connectivity. As can be seen in Figures~\ref{fig:nli_interface} and ~\ref{fig:ve_interface}, for both NLI and VE, the interface included the premise (either textual or visual), followed by the textual hypothesis.
Annotators were then asked to provide a label, a rationale for picking the label, as well as a confidence score.

\section{Qualitative Dataset Examples}
\label{sec:appendix_examples}

\begin{figure*}[t]
    \centering
    \includegraphics[width=\textwidth]{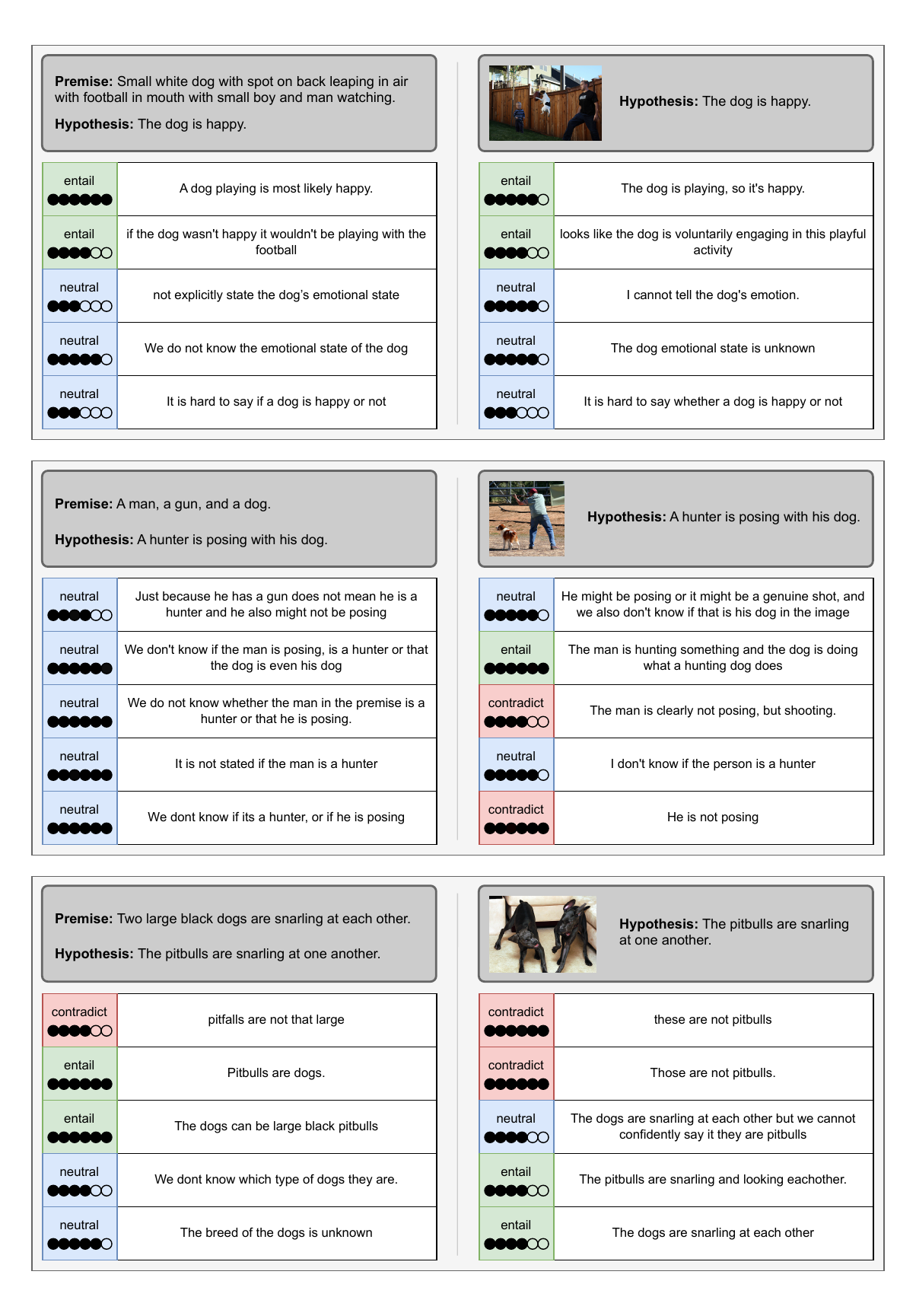}
    \caption{Three selected samples in both NLI and VE, complete with all five annotations. Each bounding box represents one sample. The left side shows the textual variant and the right side shows the visual one. Below each instance we show the five annotations collected for each sample. Annotations on the same line signify annotations given by the same annotator. Each annotation includes the label, confidence score, and explanation.}
    \label{fig:qual_examples}
\end{figure*}

We additionally conduct a small-scale exploratory qualitative analysis on selected samples, three of which can be found in Figure~\ref{fig:qual_examples}. Each bounding box represents one sample, where its left side shows the textual variant and the right side shows the visual one. Below each instance are the five annotations collected for it. To facilitate by-annotator comparisons across NLI and VE, annotations on the same line signify annotations given by the same annotator. Each annotation includes the label, confidence score, and explanation.

In the first example, both the textual and visual premise present a dog playing with a football, while the hypothesis to be judged states that \emph{the dog is happy}.
In both NLI and VE, the sample received both neutral ($\frac{3}{5}$) and entailment labels ($\frac{2}{5}$). Interestingly, individual annotators retained their labels across the two setups: annotators voting neutral in the NLI version also voted neutral in the VE version, and the same applies to the entailment label.
Looking at their explanations, annotators selecting entailment placed their vote on the assumption that \emph{a dog playing is most likely happy}, while the ones that voted neutral rather focused on the fact that \emph{It is hard to say whether a dog is happy or not}. As such, this instance seems to represent a particular subset of instances that we encountered in both NLI and VE samples: in these instances, the concept of interest is genuinely up for interpretation; e.g., how can one tell the emotional state of an animal?

The second example shows how an originally unanimously annotated sample in the textual variant ($\frac{5}{5}$ neutral) can turn into a more contentious one in the visual version ($\frac{2}{5}$ neutral, $\frac{2}{5}$ contradiction, and $\frac{1}{5}$ entailment). The hypothesis states that \emph{A hunter is posing with his dog}, which, in combination with the NLI premise \emph{A man, a gun, and a dog} results in a neutral label, with annotators pointing out not knowing whether the man is a hunter, whether he is posing, or whether the dog is truly his dog.
This changes when the textual premise is replaced by the image, representing another kind of subset we observed: here, rather than taking in the whole image all at once, annotators seem to ''zoom in'' and focus only on very specific aspects of it, and place their judgments solely based on this restricted view.
In this instance, the annotator choosing entailment states that \emph{The man is hunting something and the dog is doing what a hunting dog does}, focusing on the aspect of the man being a hunter, while passing over the posing aspect (i.e., while the man could very well be a hunter, that does not necessarily mean that he is posing). Annotators selecting contradiction, on the other hand, focus solely on the posing aspect, stating that, to them, \emph{The man is clearly not posing, but shooting}.
While the remainder of annotators chose a neutral label, raising similar questions and uncertainties as in the NLI version, it seems that some people may employ a cognitive simplification strategy here: when the interpretive space of the image feels too open, annotators may simplify it by focusing on specific aspects rather than ruminating over everything that could or could not be happening within or beyond the frame.

The third example is the only instance that did not converge to a single majority label in both modalities (NLI: $\frac{2}{5}$ neutral, $\frac{2}{5}$ entailment, and $\frac{1}{5}$ contradiction, VE: $\frac{2}{5}$ entailment, $\frac{2}{5}$ contradiction, and $\frac{1}{5}$ neutral). The textual and visual premises present \emph{Two large black dogs are snarling at each other} and the hypothesis states that \emph{The pitbulls are snarling at one another}. Interestingly, while the image and text seem quite aligned in terms of posited information, the different modes of presenting it seem to influence annotator judgments. Notably, in the visual variant, interpretation is particularly contingent on annotators' knowledge about dog breeds and their appearance, with some annotators accepting the dogs present in the image as pitbulls, others stating that they \emph{[...] are not pitbulls}, whereas others state that they \emph{[...] cannot confidently say if they are pitbulls}.

\section{Prompt Templates}
\label{sec:appendix_prompting}

The prompt templates used for the NLI and VE task are shown in Figure \ref{fig:nli_prompt} and Figure \ref{fig:ve_prompt}, respectively. To avoid option and position bias \citep{zheng_ICLR2024_mc_robust, pezeshkpour-hruschka-2024-large}, we independently randomize, for each instance, the mapping between answer option letters \texttt{\{A, B, C\}} and labels \textit{entailment}, \textit{neutral}, and \textit{contradiction}, together with their corresponding definitions.

\begin{figure}[h!]
\centering
\begin{tcolorbox}[width=\columnwidth]

\textbf{Prompt:}

You are given a premise and a hypothesis. Determine whether the premise entails, contradicts, or is neutral with respect to the hypothesis.

\medskip

\textbf{Definitions:}

\smallskip

\textbf{A. entailment:} The hypothesis is definitely true given the premise.

\textbf{B. contradiction:} The hypothesis is definitely false given the premise.

\textbf{C. neutral:} The premise does not provide enough information to determine whether the hypothesis is true or false.

\medskip

\textbf{Premise:}
`\{\texttt{premise}\}'

\medskip
\textbf{Hypothesis:}
`\{\texttt{hypothesis}\}'

\medskip

You only need to output the option letter, no explanation.

\medskip

\textbf{Answer:}
\end{tcolorbox}
\caption{Prompt template used for natural language inference.}
\label{fig:nli_prompt}
\end{figure}

\begin{figure}[h!]
\centering
\begin{tcolorbox}[width=\columnwidth]

\texttt{<image>}

\medskip

\textbf{Prompt:}

You are given an image and a hypothesis. Determine whether the image entails, contradicts, or is neutral with respect to the hypothesis.

\medskip

\textbf{Definitions:}

\smallskip

\textbf{A. entailment:} The hypothesis is definitely true given the image.

\textbf{B. contradiction:} The hypothesis is definitely false given the image.

\textbf{C. neutral:} The image does not provide enough information to determine whether the hypothesis is true or false.

\medskip

\textbf{Hypothesis:}
`\{\texttt{hypothesis}\}'

\medskip

You only need to output the option letter, no explanation.

\medskip

\textbf{Answer:}
\end{tcolorbox}
\caption{Prompt template used for visual entailment.}
\label{fig:ve_prompt}
\end{figure}

\section{Details on Evaluation Metrics}
\label{sec:appendix_metrics}

\paragraph{Human Label Entropy.} We quantify annotator disagreement within NLI and VE using Shannon entropy \citep{shannon1948entropy} over the empirical human label distribution.
For each example $i$ and modality $m \in \{\mathrm{NLI}, \mathrm{VE}\}$, we obtain a distribution $p_i^m$ by normalizing the counts of the five annotations over the label set $\mathcal{Y} = \{\textsc{ent}, \textsc{neu}, \textsc{con}\}$.
We compute human label entropy as:

\[
H(p_i^m) = - \sum_{y \in \mathcal{Y}} p_i^m(y) \log_2 p_i^m(y).
\]

Entropy is $0$ when all annotators select the same label and increases as votes are distributed across labels.
With five annotators and three labels, the maximum attainable entropy occurs for a $2$--$2$--$1$ vote split and is approximately $1.522$.
We do not further normalize entropy values.

\paragraph{Jensen-Shannon Distance.}
We use Jensen-Shannon distance \citep{endres2003jsd} to compare label distributions. Unlike KL divergence \citep{kullback1951kl}, it is symmetric and ranges from $0$ to $1$, with larger values indicating greater distributional differences. In the human analysis, we compute it between NLI and VE label distributions for each paired example.
For model analyses, we compute it between the model prediction distribution and the corresponding human label distribution.
Given two distributions $p$ and $q$ over $\mathcal{Y}$ and their average $m = \frac{1}{2}(p + q)$, we compute: \[
\mathrm{JSD}(p, q)
=
\sqrt{
\frac{1}{2}D_{\mathrm{KL}}(p \| m)
+
\frac{1}{2}D_{\mathrm{KL}}(q \| m)
},
\]
where the KL divergence is defined as:
\[
D_{\mathrm{KL}}(a \| b)
=
\sum_{y \in \mathcal{Y}} a(y)\log_2 \frac{a(y)}{b(y)}.
\]

\paragraph{Total Variation Distance.}

In addition to JSD, we use TVD \citep{devroye_tvd_2001} to compare label distributions.
Unlike JSD, TVD is based on absolute differences between probability masses and can be interpreted as the amount of probability mass that has to be shifted to transform one distribution into the other.
As with JSD, we compute TVD to both compare human label distributions across NLI and VE, and to compare human label distributions with model prediction distributions.
Given two distributions $p$ and $q$ over $\mathcal{Y}$, we compute:

\[
\mathrm{TVD}(p, q)
=
\frac{1}{2}
\sum_{y \in \mathcal{Y}}
\left|p(y) - q(y)\right|.
\]

Like JSD, TVD also ranges from $0$ to $1$, with higher values indicating greater differences in distributions.

\paragraph{Normalized Predictive Entropy.}
We use normalized predictive entropy as a measure of model uncertainty.
For each instance $i$ and modality $m \in \{\mathrm{NLI}, \mathrm{VE}\}$, let $q_i^m$ denote the model distribution over the label set $\mathcal{Y} = \{\textsc{ent}, \textsc{neu}, \textsc{con}\}$, obtained from the answer-option logits.
We compute:

\[
H_{\mathrm{norm}}(q_i^m)
=
-\frac{1}{\log |\mathcal{Y}|}
\sum_{y \in \mathcal{Y}} q_i^m(y)\log q_i^m(y).
\]

Values close to  $0$ indicate that the model assigns most probability mass to a single label, while values close to $1$ indicate near uniform label distribution and therefore greater uncertainty.

\paragraph{Conformal Prediction with Least Ambiguous set-valued Classifier.}
In addition to normalized predictive entropy, we use conformal prediction with the Least Ambiguous set-valued Classifier (LAC), following \citet{ye_2024_uncertainty}.
Conformal prediction turns model probabilities into calibrated prediction sets with a target coverage level, providing a set-valued uncertainty estimate.
The procedure follows three steps: compute LAC scores on a held-out calibration set, estimate a conformal threshold on these scores, and use this threshold to construct prediction sets for test instances.
For a calibration instance $i$, let $q_i$ denote the model distribution over $\mathcal{Y}$ and let $y_i$ be the true label.
The LAC score is defined as:
\[
s_i = 1-q_i(y_i).
\]

Given calibration scores $s_1,\ldots,s_n$, we compute the threshold:
\[
\hat{q}
=
\operatorname{Quantile}_{\lceil (n+1)(1-\alpha)\rceil/n}
(s_1,\ldots,s_n).
\]

At test time, the prediction set for instance $i$ is given by
\[
C_i
=
\{y' \in \mathcal{Y}: 1-q_i(y') \leq \hat{q}\}.
\]
We use the resulting prediction set size as the uncertainty score, with larger sets indicating greater model uncertainty.
We set $\alpha=0.1$ ($90\%$ target coverage rate).
For calibration, we use a set of paired SNLI/e-SNLI-VE instances with their original single gold labels ($n=988$) that are disjoint from the 225 pairs used for human annotation.
\citet{ye_2024_uncertainty} consider both LAC and Adaptive Prediction Sets (APS).
In our three-label setting, APS saturates for VE: when the gold label is ranked last, its cumulative APS score is necessarily $1$, which occurs frequently enough in the VE calibration data for the $90\%$ coverage threshold to reach $1$.
Consequently, all three labels are included in every VE prediction set, leaving no variation for instance-level analyses.
We therefore use LAC, which retains variation in prediction set size in our setting.

\paragraph{Spearman's Rank Correlation.}

To inspect whether model uncertainty is aligned with human uncertainty, we compute Spearman's rank correlation $\rho$ \citep{spearman1904} between human and model uncertainty scores.
For human uncertainty, we use human label entropy and inverse confidence, defined as $7-c_i$, where $c_i$ is the mean human confidence score on the original $1$--$6$ scale.
For model uncertainty, we use normalized predictive entropy and LAC set size.
Given paired uncertainty scores $(x_i, z_i)$, where $x_i$ denotes a human uncertainty score, and $z_i$ denotes a model uncertainty score,  let $r_i = R(x_i)$ and $s_i = R(z_i)$ denote their ranks among the examples.
We compute:

\[
\rho
=
\frac{
\sum_i (r_i-\bar r)(s_i-\bar s)
}{
\sqrt{
\sum_i (r_i-\bar r)^2
\sum_i (s_i-\bar s)^2
}
}.
\]

We compute correlations separately for each model in NLI and VE.
Positive correlations indicate that instances with higher human uncertainty also tend to receive higher model uncertainty.

\section{Original vs. Human-Majority VE Labels.}
\label{appendix:label_shifts_reannotation}

\begin{figure}[th!]
    \centering
    \includegraphics[width=\columnwidth]{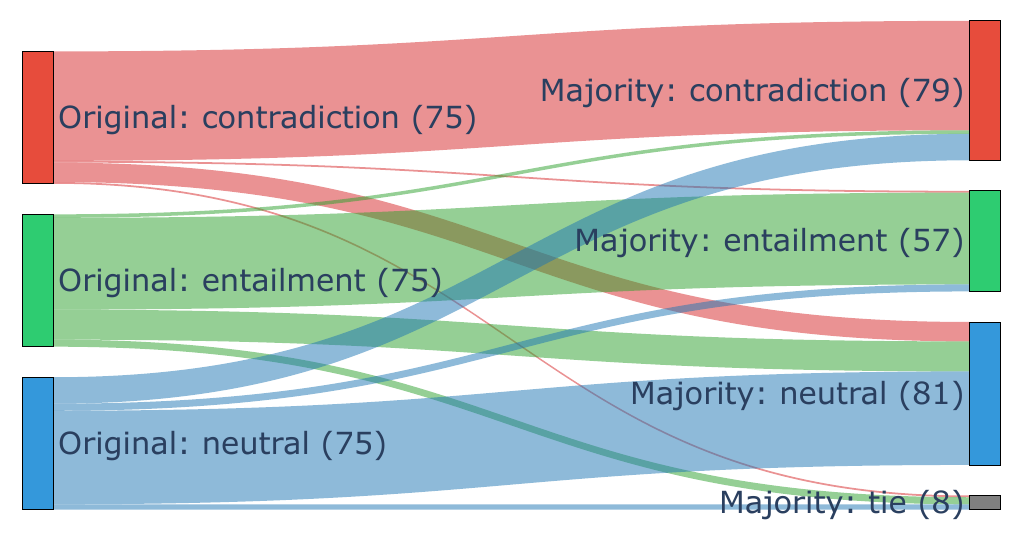}
    \caption{
    Label shifts from original single e-SNLI-VE labels to VE majority labels from our new annotations.
    }
    \label{fig:appendix_ve_reannotation}
\end{figure}

Figure~\ref{fig:appendix_ve_reannotation} compares the original single e-SNLI-VE labels with the VE majority obtained from our new annotations (5 per instance).
Most contradiction, entailment, and neutral cases retain their original label (82.67\%, 69.33\%, and 70.67\%, respectively). 
When labels change, they often move through neutral: 22.67\% of originally entailment and 14.67\% of originally contradiction cases receive a neutral majority, while direct shifts between entailment and contradiction are rare, suggesting that neutral is the main boundary area for HLV.

\section{Stability Checks for Human Analysis}
\label{appendix:stability_checks}

We assess whether our main human disagreement results are sensitive to the annotations used to estimate the label distributions. We consider two checks: removing one paired NLI--VE judgment per instance and excluding author-annotator judgments.

\paragraph{Paired leave-one-vote-out analysis.}

Because the 21 annotators labeled different subsets of instances, leaving out one annotator globally would affect only part of the dataset and produce unequal vote counts. We therefore perform a paired leave-one-vote-out analysis at the instance level. For each instance, we remove one annotator's NLI and VE judgments and recompute the metrics from the remaining four votes. We repeat this for all five possible paired removals and average the resulting values. Table \ref{tab:leave_one_vote_out} shows that results remain stable under one-vote removal.

\begin{table}[th!]
\centering
\small
\begin{tabular}{lccc}
\toprule
\textbf{Metric} & \textbf{5 votes} & \textbf{4 votes} & \textbf{Sensitivity} \\
\midrule
Entropy (NLI) & 0.468 & 0.430 & $\pm 0.026$ \\
Entropy (VE) & 0.538 & 0.495 & $\pm 0.026$ \\
$\Delta$ entropy (VE--NLI) & 0.070 & 0.065 & $\pm 0.036$ \\
TVD (VE--NLI) & 0.484 & 0.487 & $\pm 0.013$ \\
JSD (VE--NLI) & 0.524 & 0.526 & $\pm 0.013$ \\
\bottomrule
\end{tabular}
\caption{Paired leave-one-vote-out stability analysis. The 4-vote values average over all five possible paired removals per instance. Sensitivity reflects variation across these removals.}
\label{tab:leave_one_vote_out}
\end{table}

\paragraph{Author-exclusion analysis.}

Three of the 21 annotators were authors of this paper, which could introduce bias into the human disagreement results. We therefore recompute the main metrics after excluding all author-annotator judgments. Of the 225 instances, 37 (16.4\%) were annotated entirely by non-authors, 162 (72.0\%) included one author-annotator, and 26 (11.6\%) included two. Author-annotators thus never constituted a majority for any instance, and all 225 instances remain represented by three to five non-author annotators after exclusion.
We report results in Table \ref{tab:author_exclusion}.
The main findings persist after excluding author-annotators: the aggregate entropy difference between NLI and VE remains indistinguishable from zero and moves closer to zero, while TVD and JSD remain nearly unchanged. Because author exclusion leaves between three and five judgments per instance, direct numerical differences between the two columns should be interpreted cautiously.

\begin{table}[th!]
\centering
\small
\setlength{\tabcolsep}{3pt}
\begin{tabular}{lcc}
\toprule
\textbf{Metric} & \textbf{All annotators} & \textbf{w/o authors} \\
\midrule
Entropy (NLI) 
& 0.468 {\scriptsize [0.41, 0.53]} 
& 0.467 {\scriptsize [0.40, 0.53]} \\

Entropy (VE) 
& 0.538 {\scriptsize [0.48, 0.60]} 
& 0.469 {\scriptsize [0.40, 0.54]} \\

$\Delta$ entropy (VE--NLI) 
& 0.070 {\scriptsize [-0.01, 0.15]} 
& 0.002 {\scriptsize [-0.08, 0.09]} \\

TVD (VE--NLI) 
& 0.484 {\scriptsize [0.44, 0.53]} 
& 0.488 {\scriptsize [0.44, 0.53]} \\

JSD (VE--NLI) 
& 0.524 {\scriptsize [0.48, 0.57]} 
& 0.526 {\scriptsize [0.48, 0.57]} \\
\bottomrule
\end{tabular}
\caption{Human disagreement metrics with all annotators and after excluding author-annotators. Values are means with 95\% percentile bootstrap confidence intervals based on 10,000 resamples of the 225 instances.}
\label{tab:author_exclusion}
\end{table}

\section{Examples of Label--Rationale Patterns}
\label{sec:appendix_explanation_examples}

\begin{figure*}[th!]
    \centering
    \includegraphics[width=\textwidth]{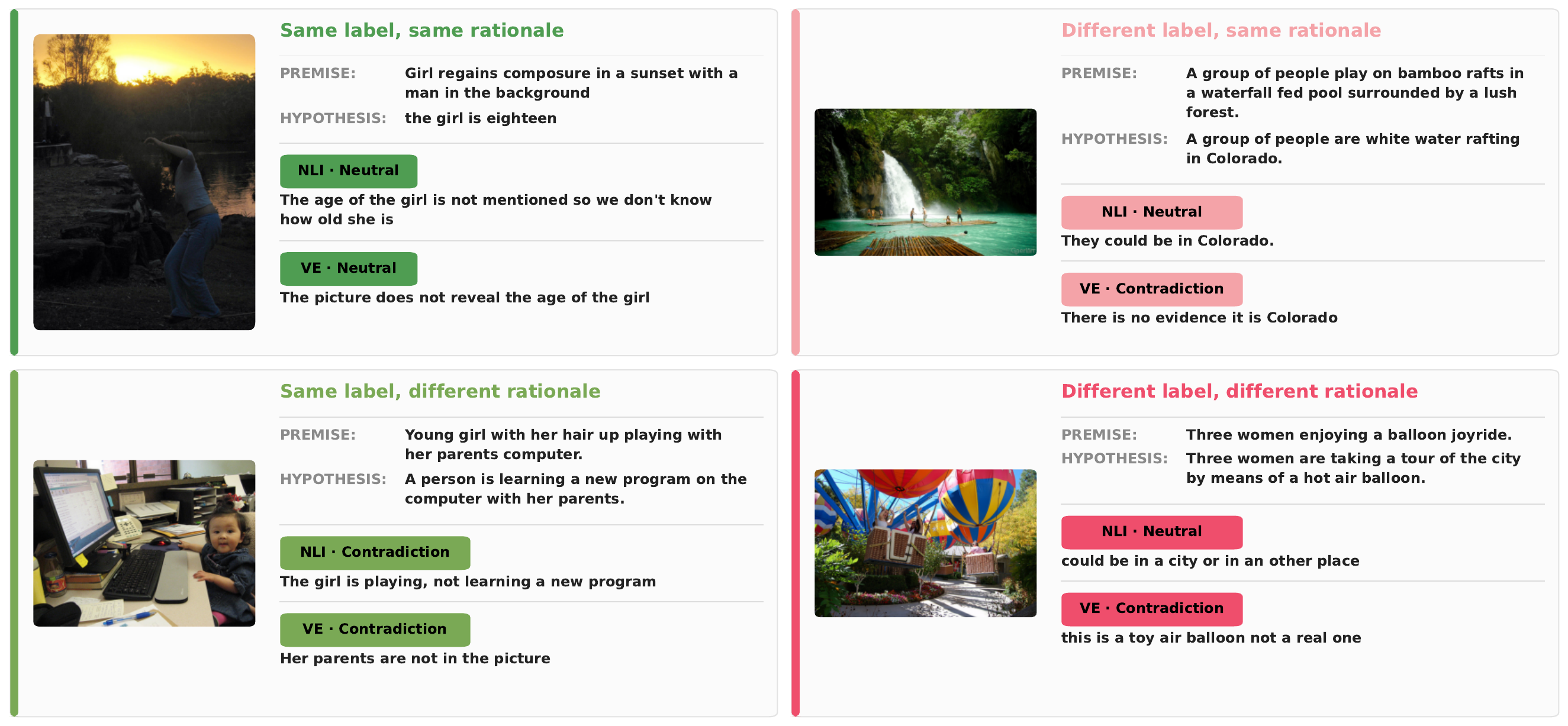}
    \caption{Illustrative examples of the four label-rationale patterns used in the explanation analysis. Each panel shows the textual premise, hypothesis, and paired NLI and VE judgments from the same annotator.}
    \label{fig:appendix_label_rationale_examples}
\end{figure*}

Figure~\ref{fig:appendix_label_rationale_examples} provides one illustrative example for each of the four label-rationale patterns used in our explanation analysis.
We consider two rationales the same when they rely on the same underlying consideration, even if their wording differs.

\paragraph{Same label, same rationale.}
In this example, the annotator assigns neutral in both NLI and VE.
In both explanations, the judgment rests on the fact that the girl's age cannot be inferred from the available evidence.

\paragraph{Same label, different rationale.}
In this case, the annotator assigns contradiction in both NLI and VE, but for different reasons.
The NLI explanation focuses on the girl playing rather than learning a new program, whereas the VE explanation instead points to the absence of her parents from the image.

\paragraph{Different label, same rationale.}
Here, the label changes from neutral to contradiction, but both explanations hinge on whether the scene can be established as being in Colorado.

This pattern is present in 14 out of 850 NLI-VE explanation pairs. While it might seem counterintuitive at first, it does not necessarily signify an annotation error; it may also be indicative of individual differences in label interpretation. For instance, in the present example, the annotator seems to interpret neutral as allowing for the possibility that the scene is in Colorado (\emph{They could be in Colorado}), while treating the absence of evidence that it is in Colorado as sufficient for contradiction (\emph{There is no evidence it is Colorado}).
While the annotator thus deviates from the original label interpretation,
where such uncertainty would generally fall under the neutral label, \citet{hong-etal-2026-agree} report comparable cases in which annotators provide similar reasoning but ultimately choose different labels, suggesting a degree of subjectivity in label boundaries.

\paragraph{Different label, different rationale.}
In this example, the judgment changes from neutral to contradiction, and the rationale changes as well.
The NLI explanation leaves open whether the balloon ride is a city tour, while the VE explanation rejects that possibility.

\section{Additional Model Results}

\subsection{Text-only LLM Results}
\label{sec:appendix_text_only_llms}

Table \ref{tab:qwen3_text_only} reports results for text-only Qwen3 models on NLI. These models serve as a reference point for interpreting VLM performance on the text-only task. The smallest LLM, Qwen3-1.7B, underperforms all VLMs in terms of accuracy, TVD and JSD, while Qwen3-4B and and Qwen3-8B fall within the range of the VLMs on all three metrics, although they remain below the strongest VLMs.
This suggests that strong text-only LLMs and VLMs perform comparably on the NLI task.
The uncertainty metrics (normalized predictive entropy and LAC set size) reveal a similar pattern, with values that are broadly comparable to the VLM uncertainty scores on NLI.

\begin{table}[h!]
\centering
\small
\setlength{\tabcolsep}{4pt}
\begin{tabular}{l ccc cc}
\toprule
\textbf{Model} 
& \textbf{Acc. $\uparrow$} 
& \textbf{TVD $\downarrow$} 
& \textbf{JSD $\downarrow$}
& \textbf{Ent.}
& \textbf{LAC} \\
\midrule
Qwen3-1.7B & 42.1 & 0.556 & 0.583 & 0.167 & 2.77 \\
Qwen3-4B   & 81.4 & 0.259 & 0.304 & 0.021 & 1.74 \\
Qwen3-8B   & 83.7 & 0.249 & 0.297 & 0.056 & 2.04 \\
\bottomrule
\end{tabular}
\caption{
Text-only Qwen3 results on NLI. Ent. denotes normalized predictive entropy, and LAC denotes conformal prediction set size using LAC score.
}
\label{tab:qwen3_text_only}
\end{table}

\subsection{Accuracy on Neutral Instances Across Modalities}
\label{sec:appendix_neutral_accuracies}

Table~\ref{tab:appendix_ve_neutral_accuracy} reports accuracy on the 81 paired instances whose VE majority label is neutral.
In the full paired evaluation set, the NLI examples are mostly neutral, while the corresponding VE examples are more evenly distributed across entailment, neutral and contradiction.
To check whether the VE drop persists when this label-distribution difference is removed, we restrict the comparison the the instances that have a neutral majority label in both modalities.
Even under this restriction, all models obtain lower accuracy in VE than NLI.

\begin{table}[th!]
\centering
\small
\setlength{\tabcolsep}{4pt}
\begin{tabular}{lcc}
\toprule
\textbf{Model} & \textbf{NLI acc.} & \textbf{VE acc.} \\
\midrule
Gemma-3-4B  & 76.2 & \ws{21}{55.6} \\
Gemma-3-12B & 80.0 & \ws{28}{51.9} \\
Gemma-3-27B & 87.5 & \ws{34}{53.1} \\

InternVL3.5-2B & 82.5 & \ws{29}{53.1} \\
InternVL3.5-4B & \underline{92.5} & \ws{23}{69.1} \\
InternVL3.5-8B & 88.8 & \ws{23}{65.4} \\

LLaVA-OneVision-4B & 91.2 & \ws{42}{49.4} \\
LLaVA-OneVision-8B & 87.5 & \ws{34}{53.1} \\

Molmo2-4B   & 88.8 & \ws{12}{76.5} \\
Molmo2-8B   & \textbf{96.2} & \ws{14}{\underline{82.7}} \\
Molmo2-O-7B & 83.8 & \ws{23}{60.5} \\

Qwen3VL-2B & 76.2 & \ws{34}{42.0} \\
Qwen3VL-8B & \textbf{96.2} & \ws{12}{\textbf{84.0}} \\
\bottomrule
\end{tabular}
\caption{
Accuracy on paired instances whose VE majority label is neutral ($n=81$). VE cells are shaded according to the decrease in accuracy compared to the corresponding NLI instances, with darker cells indicating larger drops.
}
\label{tab:appendix_ve_neutral_accuracy}
\end{table}

\subsection{Model Predictions vs Human Labels}
\label{sec:appendix_distr_align}

\begin{figure}[t]
    \centering
    \includegraphics[width=\linewidth]{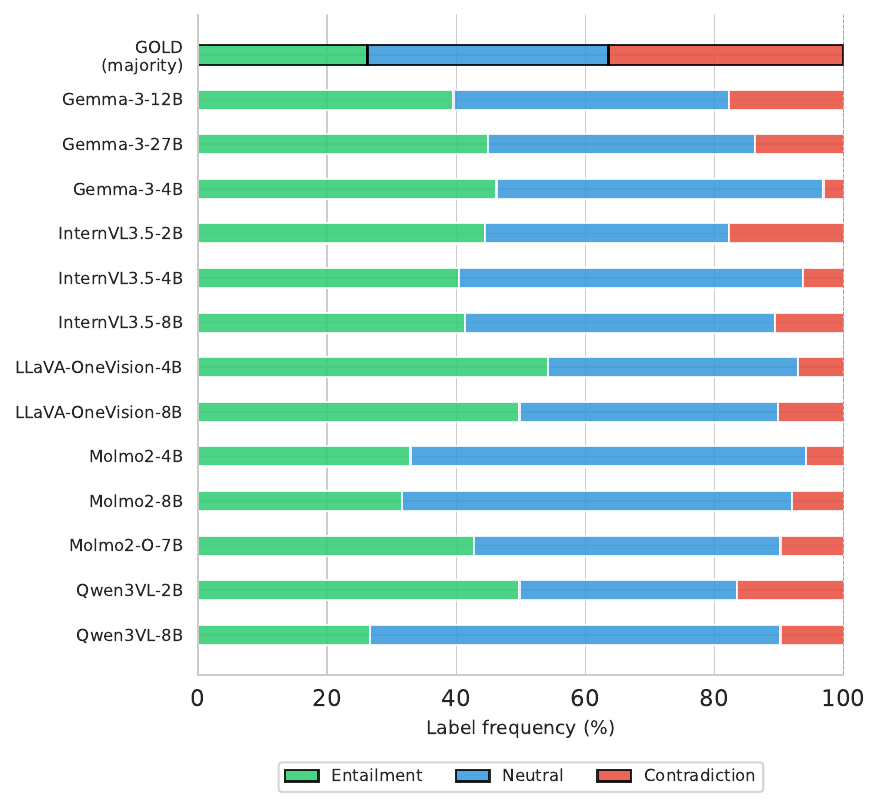}
    \caption{Predicted vs human label distribution for VE across our 225 instances.}
    \label{fig:appendix_model_label_dist}
\end{figure}

Figure~\ref{fig:appendix_model_label_dist} compares the distribution of labels predicted by the models for the VE task with the human-majority VE label distribution.
Compared to humans, models tend to underpredict the contradiction label and instead assign more predictions to entailment and neutral.

Table~\ref{tab:appendix_ve_labelwise_alignment} shows a label-wise breakdown of human-model distributional alignment in VE.
For each label, we compare the model judgment distribution with the corresponding human label distribution using TVD and JSD.
Alignment is strongest for entailment and weaker for neutral, and especially low for contradiction cases.
This mirrors the aggregate label-distribution pattern where models tend to underpredict the contradiction label.

\subsection{Model Uncertainty Analyses}
\label{sec:appendix_distr_align}

\paragraph{Detailed Model Uncertainty Results.}
\label{sec:appendix_full_uncertainty}

Table~\ref{tab:appendix_model_uncertainty_full} reports detailed model uncertainty scores for NLI and VE.
We include normalized predictive entropy and conformal prediction with LAC set size, where larger values indicate greater model uncertainty.
Normalized predictive entropy varies considerably across models, but models are more uncertain in VE than NLI in most cases.
LAC set size shows a similar trend, with larger prediction sets for VE across models.

\begin{table}[th!]
\centering
\small
\setlength{\tabcolsep}{4pt}
\begin{tabular}{lcc cc}
\toprule
\multirow{2}{*}{\textbf{Model}}
& \multicolumn{2}{c}{\textbf{Entropy}}
& \multicolumn{2}{c}{\textbf{LAC}} \\
\cmidrule(lr){2-3} \cmidrule(lr){4-5}
& \textbf{NLI} & \textbf{VE}
& \textbf{NLI} & \textbf{VE} \\
\midrule
Gemma-3-4B
& 0.045 & 0.041
& 1.476 & \dcell{100}{2.627} \\

Gemma-3-12B
& 0.025 & \dcell{31}{0.105}
& 1.716 & \dcell{42}{2.200} \\

Gemma-3-27B
& 0.010 & \dcell{6}{0.027}
& 1.409 & \dcell{77}{2.293} \\

InternVL3.5-2B
& 0.387 & 0.346
& 2.080 & \dcell{4}{2.129} \\

InternVL3.5-4B
& 0.114 & \dcell{90}{0.344}
& 1.098 & \dcell{76}{1.969} \\

InternVL3.5-8B
& 0.116 & \dcell{96}{0.361}
& 1.124 & \dcell{76}{2.004} \\

LLaVA-OneVision-4B
& 0.513 & \dcell{6}{0.529}
& 1.307 & \dcell{77}{2.196} \\

LLaVA-OneVision-8B
& 0.386 & \dcell{32}{0.468}
& 1.400 & \dcell{53}{2.004} \\

Molmo2-4B
& 0.181 & \dcell{89}{0.407}
& 1.111 & \dcell{78}{2.004} \\

Molmo2-8B
& 0.089 & \dcell{97}{0.338}
& 1.000 & \dcell{100}{2.151} \\

Molmo2-O-7B
& 0.229 & \dcell{58}{0.377}
& 1.431 & \dcell{75}{2.289} \\

Qwen3VL-2B
& 0.096 & \dcell{58}{0.245}
& 1.907 & \dcell{31}{2.258} \\

Qwen3VL-8B
& 0.027 & \dcell{50}{0.154}
& 1.080 & \dcell{84}{2.044} \\
\bottomrule
\end{tabular}%
\caption{
Detailed model uncertainty scores measured with normalized predictive entropy and LAC set size.
VE cells are shaded according to their increase over the corresponding NLI value.
}
\label{tab:appendix_model_uncertainty_full}
\end{table}

\paragraph{Spearman Correlations with LAC Set Size.}

\begin{figure}[h!]
    \centering
    \includegraphics[width=\linewidth]{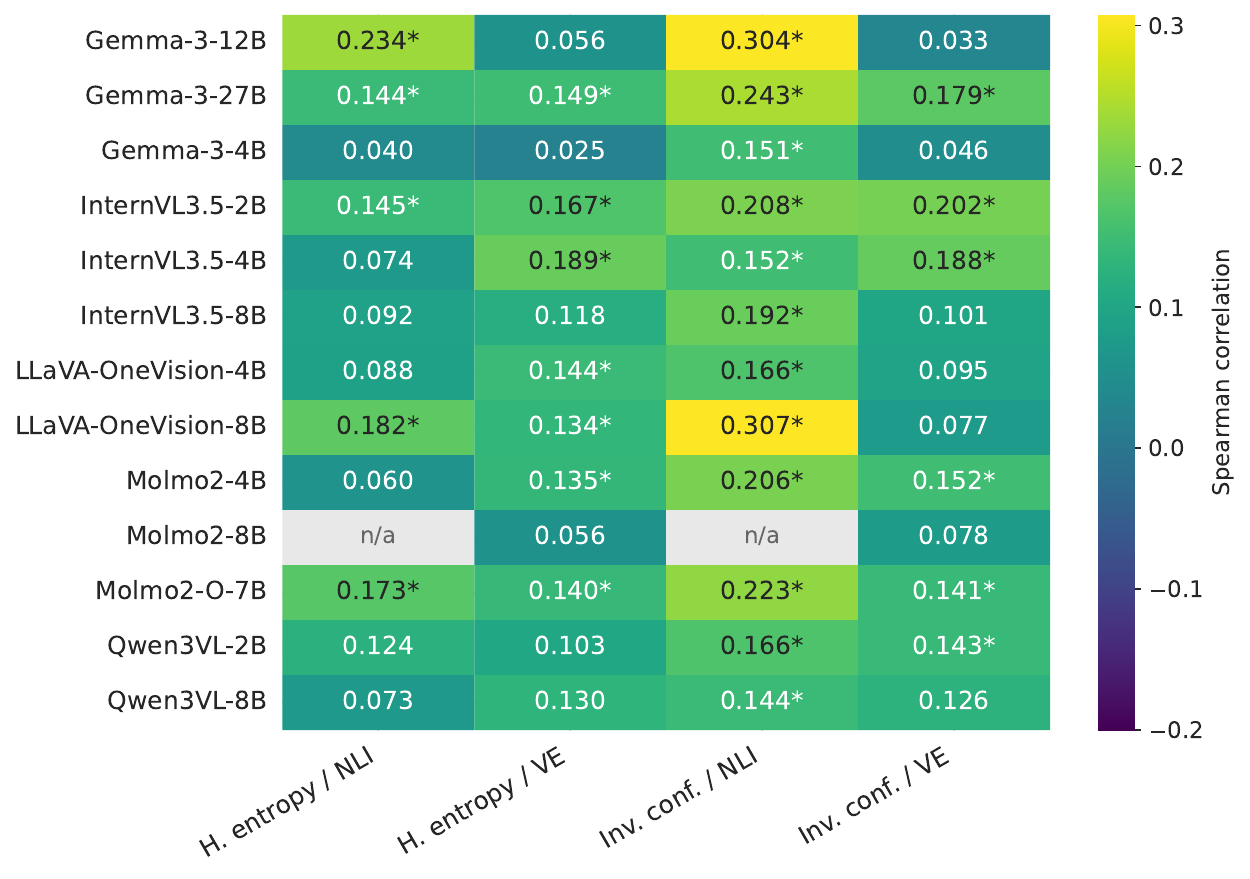}
    \caption{
    Spearman's
    $\rho$ between LAC set size and human uncertainty measures (human label entropy and inverse mean confidence). Asterisks indicate p < 0.05. n/a indicates cases where Spearman’s $\rho$ is undefined because LAC set size does not vary across instances.}
    \label{fig:appendix_uncertainty_alignment_lac}
\end{figure}

Figure \ref{fig:appendix_uncertainty_alignment_lac} shows Spearman’s rank correlation between LAC set size and the two human uncertainty measures, computed separately for NLI and VE. Because our task has only three labels, LAC set size is a relatively coarse uncertainty measure, taking only values from 1 to 3. For Molmo2-8B in NLI, the set size does not vary across instances, so Spearman’s $\rho$ cannot be computed for this case. Among the remaining correlations, 17 of 24 (71\%) are statistically significant in NLI, compared with 13 of 26 (50\%) in VE. The correlations are generally small, indicating limited correspondence between LAC set size and human uncertainty.

\section{Paired NLI-VE Confidence Intervals}
\label{appendix:cis}

We report 95\% paired bootstrap confidence intervals for NLI–VE differences in model performance and uncertainty (Table~\ref{tab:model_performance_uncertainty_cis}) and human–model alignment (Table~\ref{tab:model_alignment_cis}). Intervals are based on 10,000 bootstrap resamples of paired instances. For accuracy, we define $\Delta=\mathrm{NLI}-\mathrm{VE}$, such that positive values indicate higher accuracy in NLI. For TVD, JSD, normalized predictive entropy, and LAC set size, we define $\Delta=\mathrm{VE}-\mathrm{NLI}$, such that positive values indicate poorer alignment or greater uncertainty in VE. An interval containing zero indicates that the corresponding NLI--VE difference is not clearly distinguishable from zero under this analysis.

\begin{table*}[t]
\centering
\small
\setlength{\tabcolsep}{6pt}

\begin{tabular}{@{}lcccc@{}}
\toprule
\multirow{2}{*}{\textbf{Model}} &
\multicolumn{2}{c}{\textbf{Accuracy} ($\Delta=\mathrm{NLI}-\mathrm{VE}$)} &
\multicolumn{2}{c}{\textbf{Uncertainty} ($\Delta=\mathrm{VE}-\mathrm{NLI}$)} \\
\cmidrule(lr){2-3}
\cmidrule(lr){4-5}
&
\textbf{Overall} &
\textbf{Neutral-only} &
\textbf{Norm. entropy} &
\textbf{LAC size} \\
\midrule

Gemma-3-4B
& 26.3 [17.6, 35.0]
& 20.7 [9.9, 31.7]
& \cellcolor{gray!10}$-0.004$ [$-0.027$, 0.017]\textsuperscript{\textbf{n.s.}}
& 1.151 [1.036, 1.267] \\

Gemma-3-12B
& 24.8 [16.9, 32.5]
& 28.1 [17.3, 39.5]
& 0.080 [0.051, 0.109]
& 0.484 [0.356, 0.613] \\

Gemma-3-27B
& 27.5 [19.5, 35.5]
& 34.4 [23.5, 45.6]
& 0.016 [0.001, 0.033]
& 0.884 [0.769, 1.000] \\

InternVL3.5-2B
& 21.5 [13.4, 29.5]
& 29.4 [18.4, 40.7]
& \cellcolor{gray!10}$-0.041$ [$-0.091$, 0.010]\textsuperscript{\textbf{n.s.}}
& \cellcolor{gray!10}0.049 [$-0.098$, 0.196]\textsuperscript{\textbf{n.s.}} \\

InternVL3.5-4B
& 36.6 [29.2, 43.9]
& 23.4 [13.5, 33.3]
& 0.231 [0.184, 0.278]
& 0.871 [0.756, 0.987] \\

InternVL3.5-8B
& 31.1 [23.4, 38.6]
& 23.3 [12.1, 34.6]
& 0.245 [0.198, 0.293]
& 0.880 [0.764, 0.996] \\

LLaVA-OneVision-4B
& 33.5 [25.1, 41.7]
& 41.9 [30.9, 52.9]
& \cellcolor{gray!10}0.016 [$-0.028$, 0.059]\textsuperscript{\textbf{n.s.}}
& 0.889 [0.760, 1.013] \\

LLaVA-OneVision-8B
& 31.6 [23.9, 39.2]
& 34.4 [23.2, 45.7]
& 0.082 [0.036, 0.127]
& 0.604 [0.476, 0.733] \\

Molmo2-4B
& 34.7 [27.0, 42.4]
& 12.2 [2.5, 22.2]
& 0.227 [0.185, 0.268]
& 0.893 [0.791, 0.996] \\

Molmo2-8B
& 26.9 [19.1, 34.8]
& 13.5 [4.9, 22.2]
& 0.249 [0.212, 0.287]
& 1.151 [1.044, 1.258] \\

Molmo2-O-7B
& 28.9 [20.6, 36.9]
& 23.3 [12.3, 34.4]
& 0.148 [0.096, 0.201]
& 0.858 [0.724, 0.991] \\

Qwen3VL-2B
& 17.1 [8.1, 26.1]
& 34.3 [22.2, 46.6]
& 0.149 [0.104, 0.194]
& 0.351 [0.182, 0.520] \\

Qwen3VL-8B
& 31.0 [23.5, 38.7]
& 12.3 [3.7, 21.0]
& 0.127 [0.093, 0.162]
& 0.964 [0.849, 1.080] \\

\bottomrule
\end{tabular}

\caption{
Paired NLI--VE differences in model accuracy and uncertainty with
95\% bootstrap confidence intervals based on 10,000 paired resamples.
Cells report $\Delta$ [95\% CI]. For accuracy,
$\Delta=\mathrm{NLI}-\mathrm{VE}$ and is reported in percentage points.
For normalized predictive entropy and LAC set size,
$\Delta=\mathrm{VE}-\mathrm{NLI}$.
Non-significant differences (95\% CI includes zero) are marked with
\textsuperscript{\textbf{n.s.}} and shaded in gray.
}
\label{tab:model_performance_uncertainty_cis}
\end{table*}

\begin{table*}[t]
\centering
\small
\setlength{\tabcolsep}{7pt}

\begin{tabular}{@{}lcccc@{}}
\toprule
\multirow{2}{*}{\textbf{Model}} &
\multicolumn{2}{c}{\textbf{All instances}} &
\multicolumn{2}{c}{\textbf{Neutral-only} ($n=81$)} \\
\cmidrule(lr){2-3}
\cmidrule(lr){4-5}
&
\textbf{TVD} &
\textbf{JSD} &
\textbf{TVD} &
\textbf{JSD} \\
\midrule

Gemma-3-4B
& 0.236 [0.170, 0.303]
& 0.226 [0.163, 0.290]
& 0.218 [0.136, 0.301]
& 0.222 [0.139, 0.304] \\

Gemma-3-12B
& 0.192 [0.134, 0.249]
& 0.191 [0.135, 0.246]
& 0.235 [0.155, 0.318]
& 0.242 [0.162, 0.323] \\

Gemma-3-27B
& 0.231 [0.173, 0.290]
& 0.219 [0.160, 0.277]
& 0.288 [0.208, 0.369]
& 0.283 [0.201, 0.363] \\

InternVL3.5-2B
& 0.139 [0.088, 0.190]
& 0.112 [0.067, 0.158]
& 0.198 [0.132, 0.266]
& 0.169 [0.112, 0.227] \\

InternVL3.5-4B
& 0.222 [0.171, 0.272]
& 0.200 [0.151, 0.248]
& 0.180 [0.113, 0.246]
& 0.171 [0.104, 0.236] \\

InternVL3.5-8B
& 0.192 [0.143, 0.241]
& 0.178 [0.133, 0.223]
& 0.164 [0.101, 0.225]
& 0.159 [0.098, 0.217] \\

LLaVA-OneVision-4B
& 0.207 [0.163, 0.250]
& 0.169 [0.133, 0.205]
& 0.207 [0.156, 0.258]
& 0.161 [0.119, 0.205] \\

LLaVA-OneVision-8B
& 0.195 [0.151, 0.240]
& 0.162 [0.125, 0.200]
& 0.195 [0.137, 0.251]
& 0.158 [0.109, 0.205] \\

Molmo2-4B
& 0.216 [0.169, 0.264]
& 0.191 [0.149, 0.234]
& 0.102 [0.046, 0.156]
& 0.089 [0.036, 0.139] \\

Molmo2-8B
& 0.198 [0.147, 0.251]
& 0.185 [0.138, 0.233]
& 0.141 [0.086, 0.195]
& 0.136 [0.080, 0.192] \\

Molmo2-O-7B
& 0.186 [0.134, 0.236]
& 0.166 [0.121, 0.212]
& 0.170 [0.105, 0.234]
& 0.163 [0.104, 0.222] \\

Qwen3VL-2B
& 0.110 [0.048, 0.171]
& 0.104 [0.046, 0.162]
& 0.220 [0.133, 0.308]
& 0.208 [0.125, 0.289] \\

Qwen3VL-8B
& 0.221 [0.167, 0.277]
& 0.206 [0.152, 0.260]
& 0.148 [0.086, 0.211]
& 0.155 [0.087, 0.222] \\

\bottomrule
\end{tabular}

\caption{
Paired NLI--VE differences in human--model distributional alignment
with 95\% bootstrap confidence intervals based on 10,000 paired
resamples. Cells report $\Delta$ [95\% CI], where
$\Delta=\mathrm{VE}-\mathrm{NLI}$. The neutral-only
comparison is restricted to the 81 instances whose VE majority label
is neutral.
}
\label{tab:model_alignment_cis}
\end{table*}

\section{Implementation Details}

We use the Hugging Face implementations\footnote{\url{https://huggingface.co/models}} of all models and their default model-specific parameters.
For prediction extraction, we use a single forward pass per model and instance to obtain the next-token logits for the answer option letters.
We apply a softmax over these option-letter logits and take the argmax as the predicted label.
Predictions are therefore derived directly from answer-option logits rather than from decoded generated responses. Accordingly, we do not average over multiple sampled runs.
All models were run locally on NVIDIA A100 GPUs.

\begin{table*}[t]
\centering
\small
\setlength{\tabcolsep}{4pt}
\begin{tabular}{l ccc | ccc}
\toprule
\multirow{2}{*}{\textbf{Model}}
& \multicolumn{3}{c|}{\textbf{TVD $\downarrow$}}
& \multicolumn{3}{c}{\textbf{JSD $\downarrow$}} \\
\cmidrule(lr){2-4} \cmidrule(lr){5-7}
& \textbf{Ent} & \textbf{Neu} & \textbf{Con}
& \textbf{Ent} & \textbf{Neu} & \textbf{Con} \\
\midrule

Gemma-3-4B         
& \dcell{23}{0.228} & \dcell{50}{0.497} & \dcell{86}{0.860}
& \dcell{29}{0.290} & \dcell{54}{0.541} & \dcell{86}{0.859} \\

Gemma-3-12B        
& \dcell{20}{0.198} & \dcell{48}{0.481} & \dcell{61}{0.613}
& \dcell{26}{0.260} & \dcell{53}{0.528} & \dcell{63}{0.632} \\

Gemma-3-27B        
& \dcell{22}{0.219} & \dcell{49}{0.492} & \dcell{63}{0.630}
& \dcell{27}{0.265} & \dcell{53}{0.531} & \dcell{64}{0.642} \\

InternVL3.5-2B     
& \dcell{14}{0.139} & \dcell{45}{0.454} & \dcell{51}{\textbf{0.506}}
& \dcell{20}{\textbf{0.195}} & \dcell{50}{0.500} & \dcell{54}{\textbf{0.536}} \\

InternVL3.5-4B     
& \dcell{16}{0.161} & \dcell{33}{0.329} & \dcell{67}{0.672}
& \dcell{22}{0.222} & \dcell{38}{0.382} & \dcell{67}{0.672} \\

InternVL3.5-8B     
& \dcell{14}{0.142} & \dcell{34}{0.341} & \dcell{61}{0.608}
& \dcell{21}{0.206} & \dcell{39}{0.392} & \dcell{62}{0.619} \\

LLaVA-OneVision-4B 
& \dcell{12}{\textbf{0.121}} & \dcell{43}{0.432} & \dcell{66}{0.656}
& \dcell{21}{0.209} & \dcell{47}{0.474} & \dcell{66}{0.662} \\

LLaVA-OneVision-8B 
& \dcell{14}{\underline{0.135}} & \dcell{40}{0.398} & \dcell{60}{0.597}
& \dcell{22}{0.221} & \dcell{44}{0.439} & \dcell{61}{\underline{0.607}} \\

Molmo2-4B          
& \dcell{18}{0.178} & \dcell{29}{\underline{0.290}} & \dcell{69}{0.690}
& \dcell{25}{0.250} & \dcell{35}{\textbf{0.346}} & \dcell{69}{0.687} \\

Molmo2-8B          
& \dcell{18}{0.179} & \dcell{29}{\textbf{0.287}} & \dcell{67}{0.671}
& \dcell{26}{0.257} & \dcell{35}{\underline{0.349}} & \dcell{68}{0.680} \\

Molmo2-O-7B        
& \dcell{14}{0.141} & \dcell{39}{0.388} & \dcell{63}{0.627}
& \dcell{23}{0.226} & \dcell{44}{0.441} & \dcell{64}{0.636} \\

Qwen3VL-2B         
& \dcell{14}{0.142} & \dcell{52}{0.517} & \dcell{59}{\underline{0.590}}
& \dcell{20}{\underline{0.203}} & \dcell{55}{0.551} & \dcell{61}{0.607} \\

Qwen3VL-8B         
& \dcell{22}{0.224} & \dcell{29}{0.291} & \dcell{68}{0.677}
& \dcell{26}{0.265} & \dcell{35}{0.352} & \dcell{69}{0.686} \\

\bottomrule
\end{tabular}
\caption{
VE label-wise alignment between model predictive distributions and human judgment distributions. Results are grouped by VE majority label. Lower TVD and JSD indicate closer alignment. Cells are shaded according to the distance value, with darker cells indicating weaker alignment.
}
\label{tab:appendix_ve_labelwise_alignment}
\end{table*}

\end{document}